\documentclass[10pt,twocolumn,letterpaper]{article}

\usepackage{cvpr}
\usepackage{times}
\usepackage{epsfig}
\usepackage{graphicx}
\usepackage{amsmath}
\usepackage{amssymb}
\usepackage{subfigure}
\usepackage{eqnarray}
\usepackage{gensymb}
\usepackage{multirow}
\usepackage{enumitem}
\newenvironment{tight_itemize}{
\begin{itemize}[leftmargin=20pt]
  \setlength{\topsep}{0pt}
  \setlength{\itemsep}{0pt}
  \setlength{\parskip}{0pt}
  \setlength{\parsep}{0pt}
}{\end{itemize}}

\DeclareMathOperator*{\argmin}{\arg\min}

\usepackage[pagebackref=true,breaklinks=true,letterpaper=true,colorlinks,bookmarks=false]{hyperref}

\cvprfinalcopy 

\def\cvprPaperID{623} 

\ifcvprfinal\fi
\begin{document}

\title{UV-GAN: Adversarial Facial UV Map Completion for Pose-invariant Face Recognition}

\author{Jiankang Deng, Shiyang Cheng, Niannan Xue, Yuxiang Zhou, Stefanos Zafeiriou\\
Imperial College London\\
{\tt\small j.deng16,
shiyang.cheng11,n.xue15,yuxiang.zhou10,s.zafeiriou@imperial.ac.uk}
}

\maketitle

\begin{abstract}

Recently proposed robust 3D face alignment methods establish either dense or sparse correspondence between a 3D face model and a 2D facial image. The use of these methods presents new challenges as well as opportunities for facial texture analysis. In particular, by sampling the image using the fitted model, a facial UV can be created. Unfortunately, due to self-occlusion, such a UV map is always incomplete. In this paper, we propose a framework for training Deep Convolutional Neural Network (DCNN) to complete the facial UV map extracted from in-the-wild images. To this end, we first gather complete UV maps by fitting a 3D Morphable Model (3DMM) to various multiview image and video datasets, as well as leveraging on a new 3D dataset with over 3,000 identities. Second, we devise a meticulously designed architecture that combines local and global adversarial DCNNs to learn an identity-preserving facial UV completion model. We demonstrate that by attaching the completed UV to the fitted mesh and generating instances of arbitrary poses, we can increase pose variations for training deep face recognition/verification models, and minimise pose discrepancy during testing, which lead to better performance. Experiments on both controlled and in-the-wild UV datasets prove the effectiveness of our adversarial UV completion model. We achieve state-of-the-art verification accuracy, $94.05\%$, under the CFP frontal-profile protocol only by combining pose augmentation during training and pose discrepancy reduction during testing. We will release the first in-the-wild UV dataset (we refer as WildUV) that comprises of complete facial UV maps from 1,892 identities for research purposes.

\end{abstract}

\section{Introduction}

\begin{figure}
\centering
\includegraphics[width=1\linewidth]{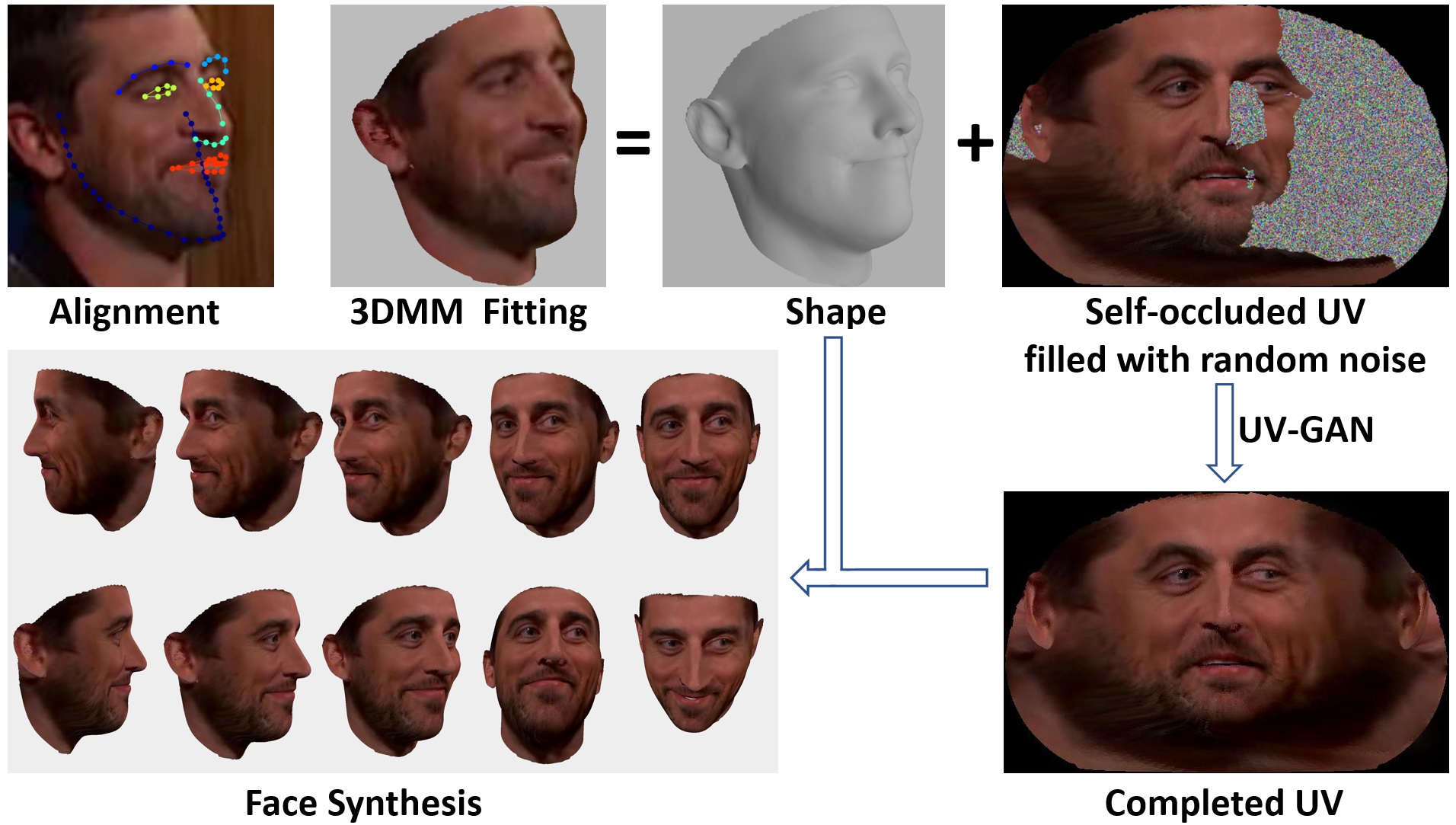}
\caption{Adversarial UV completion. After fitting a 3DMM to the image, we retrieve a 3D face with an incomplete UV map. We learn a generative model to recover the self-occluded regions. By rotating a 3D shape with complete UV map, we can generate 2D faces of arbitrary poses, which can either augment pose variations during training or narrow pose discrepancy during testing for pose-invariant face recognition.}
\vspace{-2mm}
\label{fig:UVCompletion}
\end{figure}

During the past few years, we have witnessed considerable progress in sparse and dense 3D face alignment. Some of the developments include the use of Deep Neural Networks (DNNs) to recover 3D facial structure~\cite{richardson2016learning,zhu2016face,kim2017inversefacenet}, as well as a robust framework for fitting a 3D Morphable Model (3DMM) to images in-the-wild~\cite{booth20173d}. Furthermore, benchmarks suitable for training sparse 3D face alignment models have been developed recently~\cite{zafeiriou20173d,jeni2016first}. The utilisation of these methods introduces new challenges and opportunities, as far as facial texture is concerned. More specifically, by sampling over the fitted image, a facial UV map of the texture can be created. An example of the facial UV map is shown in Fig.~\ref{fig:UVCompletion}. It is evident that such UV map contains a considerable amount of missing pixels due to self-occlusion (filled with random noise in the figure). In this paper, we tackle the problem of facial UV map completion from a single image, which has not received considerable attention. We demonstrate the usefulness of the proposed UV completion framework by creating synthetic samples to train deep neural networks for face recognition. 

Face representation using Deep Convolutional Neural Network (DCNN) embeddings is considered the method of choice for face verification, face clustering, and recognition~\cite{taigman2014deepface,schroff2015facenet,parkhi2015deep,sun2014deep}. However, when it comes to frontal-profile face verification, performance of most DCNNs drops drastically by over $10\%$~\cite{sengupta2016frontal} compared with frontal-frontal verification, while human performance only degrades slightly. This indicates that the pose variation remains a significant challenge in face recognition. One approach to learn pose-invariant discriminative representations is to collect training data with large pose variations. However, web face images usually have long-tailed pose distributions~\cite{masi2016we} and it is not feasible to collect and label data that provide good coverage for full poses for all identities. 

In this paper, we propose an adversarial UV completion framework (UV-GAN) and apply it to solve the pose-invariant face recognition problem without the requirement of extensive pose coverage in training data. Our framework is depicted in Fig.~\ref{fig:UVCompletion}, we first fit a 3DMM to 2D image and retrieve the incomplete facial UV. To complete the UV map, we combine local and global adversarial networks to learn identity-preserving full UV texture (Sec.~\ref{sect:UVcomp}). We attach the complete UV with the fitted mesh and generate synthetic data with arbitrary poses. To enlarge the diversity of training images under various poses (along with pose labels), at no additional labelling cost, we synthesise a large number of profile images from the CASIA dataset~\cite{yi2014learning}. We also use such synthesis method to reduce the pose discrepancy during testing. Our frontal-profile verification results show that rendering frontal face to profile view obtains better performance than frontalising profile face. In contrast to existing face frontalisation methods~\cite{yin2017towards,Huang_2017_ICCV}, the proposed method is more effective and flexible for pose-invariant face recognition.

To summarise, our key contributions are:
\vspace{-0.2cm}
\begin{tight_itemize}

\item We are the first to apply local, global and identity-preserving adversarial networks to the problem of UV map completion. We show that the proposed method can generate realistic and coherent UV maps under both controlled and in-the-wild settings even when the missing regions account for $50\%$ of the UV map. Using the completed UV and the corresponding 3D shape, we are able to synthesise 2D face images with arbitrary poses.

\item For face recognition training, our pose synthesis method can enrich the pose variations of training data, without incurring the expense of manually labelling large datasets spanning all poses. For recognition testing, our method can narrow the pose discrepancy between the verification pairs resulting in better performance. We obtain state-of-the-art verification accuracy, $94.05\%$, under the CFP frontal-profile protocol.

\item To the best of our knowledge, we are the first to collect a large-scale dataset of complete ear-to-ear UV facial maps of both controlled and in-the-wild data. We will release the first in-the-wild UV dataset (referred as WildUV) that comprises of complete facial UV maps from 1,892 identities for research purposes.


\end{tight_itemize}

\section{Related Work}

{\bf Image Completion} 
Image completion has been studied in numerous contexts, including inpainting and texture synthesis. Pathak~\etal~\cite{pathak2016context} propose context encoders with a reconstruction and an adversarial loss to generate the contents for missing regions that comply with the neighbourhood regions. Yang~\etal~\cite{yang2016high} further improve inpainting with a multi-scale neural patch synthesis method. This approach is based on a joint optimisation of image content and texture constraints, which not only preserves contextual structures but also produces fine details. Possible architectures, based on deep neural networks, for image completion are the so-called Pixel Recurrent Neural Networks (Pixel-RNNs) ~\cite{oord2016pixel} and Pixel CNNs~\cite{van2016conditional}. Nevertheless, we found that these architectures are mainly suitable for low-resolution images, whereas we want to design a DCNN architecture that can handle high resolutions facial UV maps. Probably the closest work to ours is the face completion method in~\cite{li2017generative}. This method combines a reconstruction loss, two adversarial losses and a semantic parsing loss to ensure genuineness and consistency of local-global contents. However, their method is trained only on a small set of pre-defined masks and is not directly applicable to our problem, since (a) each mask has semantical correspondence with a 3D face, and (b) the missing regions of profile mask may take up over $50\%$ of the image. 

{\bf Pose-invariant Face Recognition}
Recent DCNNs~\cite{taigman2014deepface,schroff2015facenet,parkhi2015deep,sun2014deep} trained on large-scale datasets have significantly boosted the performance of face recognition, the robustness of these methods against pose variations is sourced from the training data. To learn pose-invariant feature embedding, Masi~\etal~\cite{masi2016we} syntheses face image appearances across 3D viewpoints to increase pose variations in the training data. Peng~\cite{Peng_2017_ICCV} further explore reconstruction-based disentanglement during training for pose-invariant face recognition. With the introduction of GAN, pose-invariant feature disentanglement~\cite{tran2017disentangled,tran2017representation} and face frontalisation~\cite{yin2017towards,Huang_2017_ICCV} methods become quite popular. During testing, frontal faces are generated from the generator, which decreases the pose discrepancy and henceforth improves pose-invariant face recognition. However, those approaches usually rely on large amount of pairings across poses, which is over demanding under in-the-wild scenario. By contrast, we propose an adversarial UV completion method that enables us to leverage frontal and four-quarter face pairs for synthesising training examples close to the ground truth. 



\section{Proposed Approach}

\begin{figure*}
\centering
\includegraphics[width=1\textwidth]{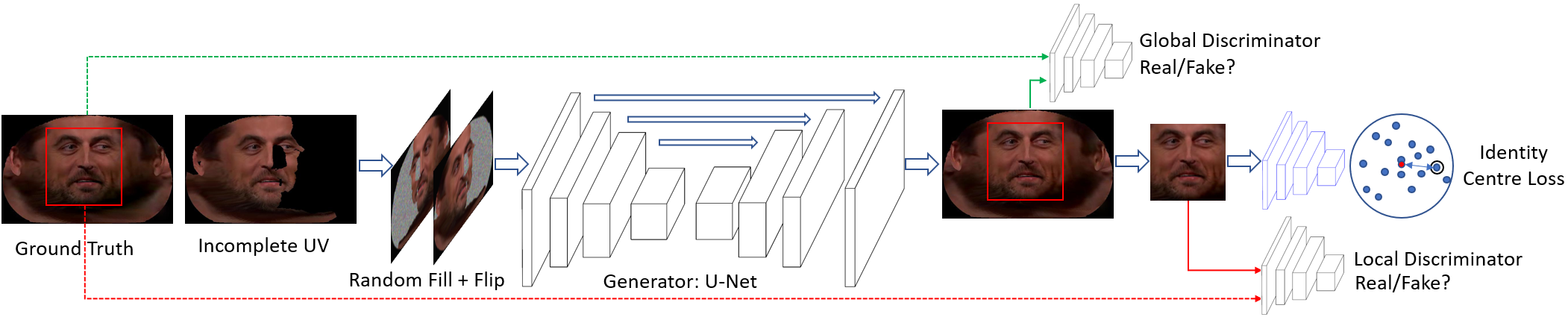}
\caption{Network architecture. It consists of one generator and two discriminators. The generator takes the incomplete UV map as input and outputs the full UV map. Two discriminators are learned to validate the genuineness of the synthetic UV texture and the main face region. The pre-trained identity classification network that would always be fixed, is to further ensure the newly generated faces preserve the identity. Note that only the generator is required for the testing stage.}
\label{fig:UV_completion_framework}
\end{figure*}

In this section, we describe the proposed model for UV completion. In brief, after a 3DMM fitting on the 2D face image, we sample the image and retrieve a UV texture for the 3D mesh. Due to face self-occlusion, such UV texture is always incomplete. Hence, our goal is to synthesise the missing contents of the UV map that are semantically consistent with the whole UV map as well as visually realistic and identity-preserved. Fig.~\ref{fig:UV_completion_framework} illustrates the proposed network consisting of one generator and two discriminators.

\subsection{3D Morphable Model Fitting}

To recover a 3D face from the 2D image, 3DMM fitting is employed. We adopt the 3DMM fitting proposed in~\cite{booth20173d} for our task. In~\cite{booth20173d}, three parametric models are defined and need to be solved: \textit{shape} (Eq.~\ref{equ:shape_model}), \textit{texture} (Eq.~\ref{equ:texture_model}) and \textit{camera} (Eq.~\ref{equ:camera_model}) models:
\vspace{-5pt}
\begin{equation}
\mathcal{S}(\mathbf{p}) = \overline{\mathbf{s}} + \mathbf{U}_s \mathbf{p},
\label{equ:shape_model}
\end{equation}
\begin{equation}
\mathcal{T}(\boldsymbol{\lambda}) = \overline{\mathbf{t}} + \mathbf{U}_t \boldsymbol{\lambda},
\label{equ:texture_model}
\end{equation}
\begin{equation}
\mathcal{W}(\mathbf{p},\mathbf{c}) = \mathcal{P}(\mathcal{S}(\mathbf{p}),\mathbf{c}),
\label{equ:camera_model}
\end{equation}
where $\mathbf{p}, \boldsymbol{\lambda}$ and $\mathbf{c}$ are shape, texture and camera parameters to optimise.
$\mathbf{U}_s$ and $\mathbf{U}_t$ are the shape and texture eigenbasis respectively, $\mathbf{\overline{s}}$ and $\mathbf{\overline{t}}$ are mean shapes of shape and texture models correspondingly, which are learnt from $10,000$ face scans of different individuals~\cite{Booth_2016_CVPR}. Function $\mathcal{P}$ is a perspective camera transformation. Thus the overall cost function for 3DMM fitting is formulated as:
\begin{equation}
\argmin_{\mathbf{p},\boldsymbol{\lambda},\mathbf{c}} 
\| \mathbf{F}(\mathcal{W}(\mathbf{p},\mathbf{c})) - \mathcal{T}(\boldsymbol{\lambda}) \|^2 + 
\alpha_s \| \mathbf{p} \|^2_{\mathbf{\Sigma}_{s}^{-1}} + \alpha_t \| \boldsymbol{\lambda} \|^2_{\mathbf{\Sigma}_{t}^{-1}}.
\label{equ:3dmm_fitting}
\end{equation}
Here, $\| \mathbf{p} \|^2_{\mathbf{\Sigma}_{s}^{-1}}$ and $\| \boldsymbol{\lambda} \|^2_{\mathbf{\Sigma}_{t}^{-1}}$ are two regularisation terms to counter over-fitting, $\mathbf{\Sigma}_{s}^{-1}$ and $\mathbf{\Sigma}_{t}^{-1}$ are diagonal matrices with the main diagonal being eigenvalues for the shape and texture model respectively. $\alpha_s$ and $\alpha_t$ are constants empirically set to weigh the two regularisation terms. Note that $\mathbf{F}(\mathcal{W}(\mathbf{p},\mathbf{c}))$ denotes the operation of sampling the feature-based input image on the projected 2D locations.

To accelerate the 3DMM fitting of the face images in the wild, we initialise the fitting with 3D landmarks\footnote{The 3D landmarks here refer to the 2D projections of the 3D facial landmarks.} predicted by~\cite{bulat2017far}, and utilise the landmarks subsequently in the optimisation by adding a 2D landmark term in Eq.~\ref{equ:3dmm_fitting}. The final objective function could be written as:
\vspace{-2mm}
\begin{align}
\argmin_{\mathbf{p},\boldsymbol{\lambda},\mathbf{c}} 
&\| \mathbf{F}(\mathcal{W}(\mathbf{p},\mathbf{c})) - \mathcal{T}(\boldsymbol{\lambda}) \|^2 + 
\alpha_l \| \mathcal{W}_l(\mathbf{p},\mathbf{c})) - \mathbf{s}_l \|^2\notag\\
& + \alpha_s \| \mathbf{p} \|^2_{\mathbf{\Sigma}_{s}^{-1}} + \alpha_t \| \boldsymbol{\lambda} \|^2_{\mathbf{\Sigma}_{t}^{-1}}, 
\label{equ:3dmm_fitting_final}
\end{align}
where $\mathbf{s}_l$ is the 2D shape, $\alpha_l$ is a constant to weigh the landmark term. Eq.~\ref{equ:3dmm_fitting_final} could be solved by the Gauss-Newton optimisation framework in~\cite{booth20173d}. 

\begin{figure*}[h!]
\begin{center}
    \includegraphics[width=\linewidth]{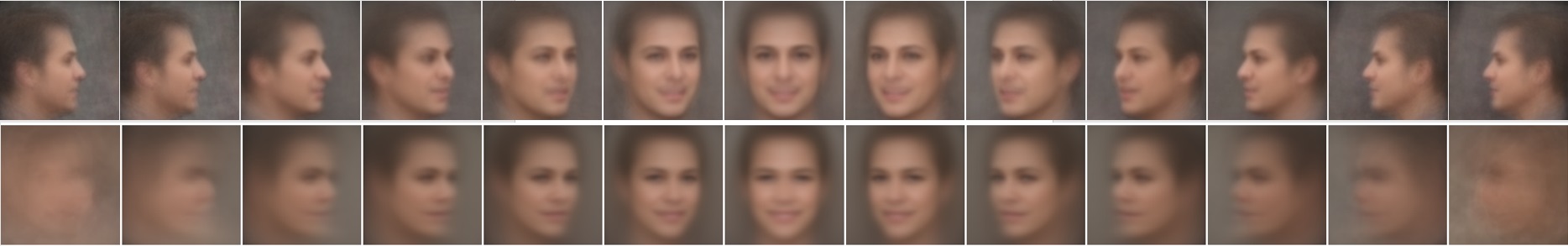}\\
\end{center}
\vspace{-4mm}
\caption{The mean faces of 13 pose groups in the CASIA dataset~\cite{yi2014learning}. The first row is our results, and the second row is from~\cite{tran2017representation}. Our mean faces are more clear under large pose variations, and there is only slight blurriness on facial organs even for profile faces.}
\label{pic:meanface}
\end{figure*}

Based on fitting results, we categorise face images from CASIA dataset~\cite{yi2014learning} into 13 pose groups~\cite{tran2017representation}. In Fig.~\ref{pic:meanface}, the mean face from~\cite{tran2017representation} becomes more blurred as the yaw angle increases. By contrast, our mean faces remain clear across all pose groups, which indicates that our fitting method produces much more accurate poses.

\subsection{UV Texture Completion}\label{sect:UVcomp}

Once we have the estimated 3D shape of the facial image, its visible vertices could be computed by z-buffering, and then utilised to generate a visibility mask for UV texture. For those \textit{invisible} parts (\ie missing textures), we want to fill them with the identity preserving textures. We propose a Generative Adversarial Network for UV completion (we refer as \textbf{UV-GAN}), it comprises of one UV generation module, two discriminators and an additional module to preserve face identity.

{\bf Generation Module} Given input UV textures with missing regions, the generator $G$ works as an auto-encoder to construct new instances. We adopt pixel-wise $l_{1}$ norm as the reconstruction loss:
\vspace{-2mm}
\begin{equation}
L_{gen}=\frac{1}{W\times H}\sum_{i=1}^{W}\sum_{j=1}^{H}\left | I_{i,j}-I_{i,j}^{*} \right |,
\label{eq:genloss}
\end{equation}
where $I_{i,j}$ is the estimated UV texture and $I_{i,j}^{*}$ is the ground truth texture. To preserve the image information in original resolution, we follow the encoder-decoder design in~\cite{isola2016image}, where skip connections between mirrored layers in the encoder and decoder stacks are made. We fill the incomplete UV texture with random noise and concatenate with its mirror image as the generator input. Since the face is not exactly symmetric, we avoid using symmetry loss as in~\cite{Huang_2017_ICCV}. Unlike the original GAN model~\cite{goodfellow2014generative} that initialises from a noise vector, the hidden representations obtained from our encoder capture more variations as well as relationships between invisible and visible regions, and thus help the decoder to fill up missing regions.

{\bf Discrimination Module} Despite that previous generation module could fill missing pixels with small reconstruction errors, it does not guarantee the output textures to be visually realistic and informative. In Fig.~\ref{fig:UVcompletionval} (a), we show examples of the generated texture from the generator, which are quite blurry and missing important details. To improve the quality of synthetic images and encourage more photo-realistic results, we adopt a discrimination module $D$ to distinguish real and fake UVs. The adversarial loss, which is a reflection of how the generator could maximally fool the discriminator and how well the discriminator could distinguish between real and fake, is defined as:
\vspace{-2mm}
\begin{align}
L_{adv} = &\mathbb{E}_{\mathbf{x}\sim p_{d}(\mathbf{x}),\mathbf{y}\sim p_{d}(\mathbf{y})}\left [ \log D(\mathbf{x,y}) \right ] + \notag\\ 
&\mathbb{E}_{\mathbf{z}\sim p_{z}(\mathbf{z}),\mathbf{y}\sim p_{d}(\mathbf{y})}\left [ 1 - \log D(G(\mathbf{z,y}),\mathbf{y}) \right ],
\label{eq:adver}
\end{align}
where $ p_{z}(\mathbf{z})$, $p_{d}(\mathbf{x})$ and $p_{d}(\mathbf{y})$  represent the distributions (\ie Gaussian distribution) of noise variables $\mathbf{z}$, full UV texture $\mathbf{x}$ and partial UV texture $\mathbf{y}$ correspondingly.

Particularly, our module $D$ consists of two discriminators: a global and a local discriminator. We firstly train a global discriminator to determine the faithfulness of the entire UV maps. The core idea is that synthetic contents should not only look realistic, but also conform to their surrounding contexts. Furthermore, we define a local discriminator that focuses on the face centre. There are two reasons for introducing the local discriminator: (1) for UV faces in the wild, outer face (\eg ears, forehead) is usually noisy and unreliable; (2) inner face is considered much more informative as fas as identity is concerned. Compared with the global discriminator, the local module (Fig.~\ref{fig:UVcompletionval} (d)) enhances the central face region with less noisy and sharper boundaries. The benefit of combining global and local discriminator is: the global one maintains the context of the facial image, while the local discriminator enforces the generated texture to be more informative within central face region. Similar GAN architecture with two discriminators could be found in~\cite{li2017generative,Huang_2017_ICCV}.

{\bf Identity Preserving Module} Preserving the identity while synthesising faces is the most critical part in developing the recognition-via-generation framework. We exploit the centre loss~\cite{wen2016discriminative} to improve the identity preserving ability of our UV-GAN. Specifically, we define the centre loss based on the activations after the average pooling layer of ResNet-27~\cite{wen2016discriminative,he2016deep}. 
\vspace{-2mm}
\begin{equation}
L_{id}=\frac{1}{m}\sum_{i=1}^{m}\left \| \mathbf{x}_i - \mathbf{c}_{y_i} \right \|_2^2,
\label{eq:centerloss}
\end{equation}
where $m$ is the batch size, $\mathbf{x}_i\in \mathbb{R}^{512}$ is the embedding features and $ \mathbf{c}_{y_i}\in \mathbb{R}^{512}$ denotes the $y_i$-th class feature centre. Note that ResNet-27 is pre-trained on CASIA~\cite{yi2014learning} dataset using the softmax loss to classify 10k identities. It captures the most prominent feature and facial structure to discriminate identity. Hence, it is beneficial to leverage this loss to maintain identity in the synthetic texture. As our feature embedding network is fixed during training, all the generated samples would lie close to their own fixed feature centre.

{\bf Objective Function} The final loss function for the proposed UV-GAN is a weighted sum of aforementioned losses:
\vspace{-2mm}
\begin{equation}
L=L_{gen}+ \lambda_{1} L_{adv\_g} + \lambda_{2}  L_{adv\_l}+ \lambda_{3} L_{id}.
\end{equation}\label{eq:loss}
$\lambda_{1}$, $\lambda_{2}$ and $\lambda_{3}$ are the weights to balance different losses.

\section{Experiments}

\subsection{Settings and Datasets}

For UV completion, the original size of ground truth UV maps is $377\times595$ and we re-scale the incomplete UV maps to $256\times256$ as the input of the UV-GAN model. The network structures of the generator and discriminator follow\footnote{https://github.com/phillipi/pix2pix}~\cite{isola2016image}. The input of the local discriminator is a $128\times128$ crop on facial feature regions as shown in Fig~\ref{fig:UV_completion_framework}, and local discriminator has the same structure as the global discriminator. For each complete facial UV map used for training, we create 20 synthetic 2D facial images with arbitrary yaw angles $[-90^{\circ},90^{\circ}]$, compute the visibility masks and use the corresponding incomplete facial UV maps for training. Our network is implemented with Tensorflow~\cite{abadi2016tensorflow}. We train UV-GAN for 100 epochs with a batch size of 8 and a learning rate of $10^{-4}$. In all our experiments, we empirically set $\lambda_{1}=10^{-2}$, $\lambda_{2}=4\times10^{-2}$, and $\lambda_{3}=10^{-3}$. 

\begin{figure}[h!]
\begin{center}
    \includegraphics[width=\columnwidth]{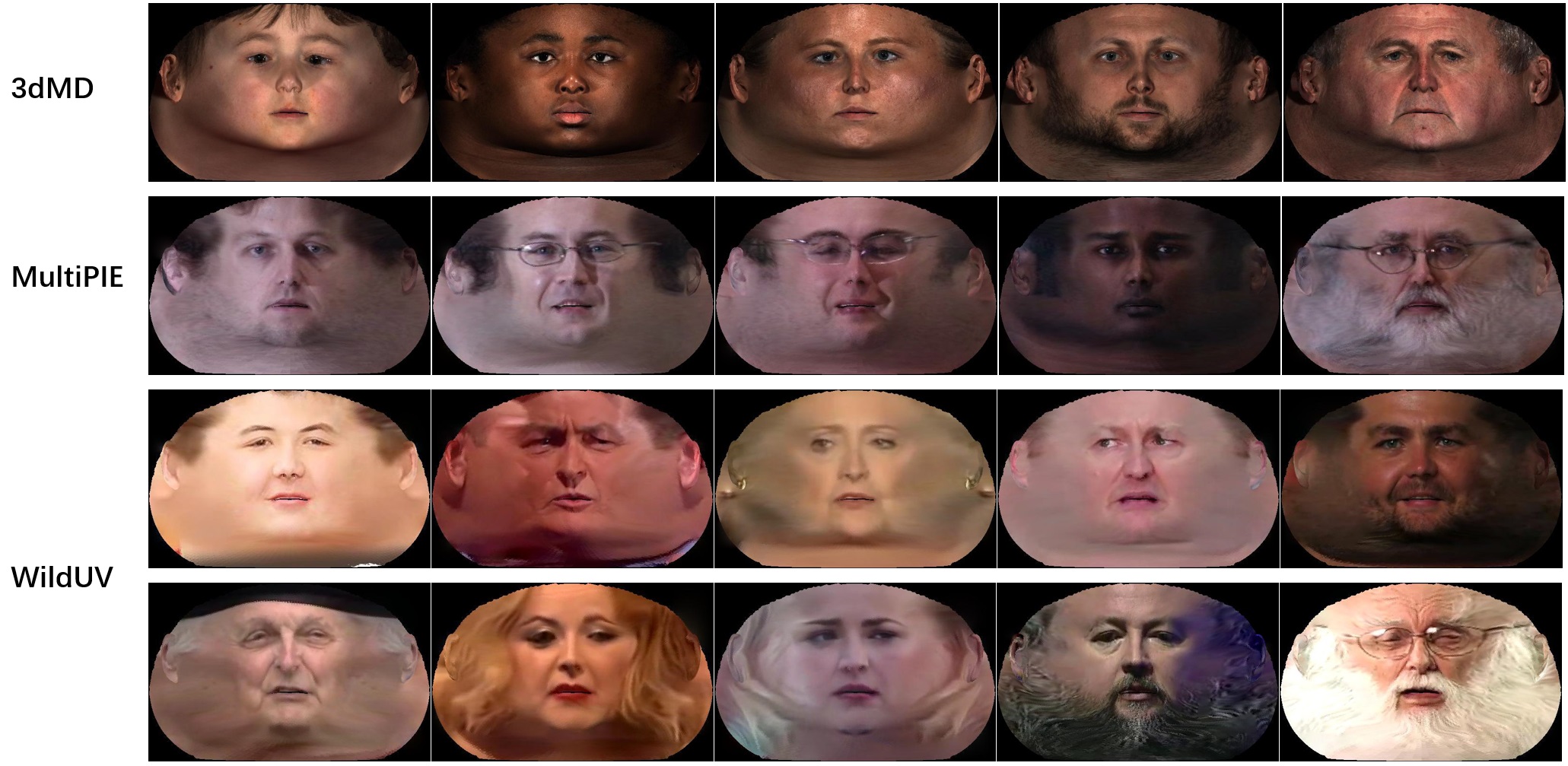}\\
\end{center}
\vspace{-4mm}
\caption{UV Completion Dataset.}
\label{pic:UVDB}
\end{figure}

{\bf UVDB}.  UVDB is a dataset that we developed for training the proposed UV-GAN. It has been built from three different sources. The first subset contains 3,564 subjects (3,564 unique identities with six expressions, 21,384 unique UV maps in total) scanned by the 3dMD device\footnote{http://www.3dmd.com/}. The second subset is constructed from Multi-PIE~\cite{gross2010multi} with 337 identities (2514 unique facial UV maps for each illumination setting, 50,280 in total). The third subset (called \textbf{WildUV}) is constructed from UMD video dataset~\cite{bansal2017dosanddonts}. We have selected videos with large pose variations to get coverage in all different poses and finally developed the first in-the-wild UV dataset with 1,892 identities (5,638 unique UV maps). Poisson blending~\cite{perez2003poisson} was used for improving the quality of the facial UV maps. As is shown in Fig.~\ref{pic:UVDB}, the UVDB contains facial UV maps captured in controlled conditions, as well as challenging in-the-wild videos (\eg, low resolution with potential occlusions). We will release WildUV subset with 2D face images, 3D fitting results and the complete UV maps. For the 3dMD and WildUV subsets, the first $90\%$ of identities are used for training and the rest are used for testing. For the MultiPIE subset, the first 200 subjects are used for training and the rest 137 subjects for testing~\cite{tran2017disentangled,tran2017representation}. For face synthesis, we use all the training subsets of UVDB.

{\bf CASIA}~\cite{yi2014learning}. CASIA dataset consists of 494,414 images of 10,575 subjects. It is a widely applied large-scale training set for face recognition. 

{\bf VGG2}~\cite{cao2017vggface2}. VGG2 dataset contains a training set with 8,631 identities (3,141,890 images) and a test set with 500 identities (169,396 images). VGG2 has large variations in pose, age, illumination, ethnicity and profession. To facilitate the evaluation of face matching across different poses, VGG2 provides face template list, which contains 368 subjects with 2 front templates, 2 three-quarter templates and 2 profile templates, with each template containing 5 images. 

{\bf MS1M}~\cite{guo2016ms}. MS1M dataset contains about 100k identities with 10 million images. We use a refined version~\cite{deng2017marginal} of MS1M to train the recognition model.

{\bf CFP}~\cite{sengupta2016frontal}. CFP dataset consists of 500 subjects, each with 10 frontal and 4 profile images. The evaluation protocol includes frontal-frontal (FF) and frontal-profile (FP) face verification, each having 10 folders with 350 same-person pairs and 350 different-person pairs.

\subsection{UV Completion}

To quantitatively evaluate the UV completion results, we employ two metrics. The first one is the peak signal-to-noise ratio (PSNR) which directly measures the difference in pixel values. The second one is the structural similarity index (SSIM)~\cite{wang2004image} that estimates the holistic similarity between two images. These two metrics are computed between the predicted UV maps and the ground truth.

\begin{figure*}[h!]
\begin{center}
\begin{tabular} {@{}c@{\hskip 0.1mm}c@{\hskip 0.1mm}c@{\hskip 0.1mm}c@{\hskip 0.1mm}c@{\hskip 0.1mm}c@{\hskip 0.1mm}c@{\hskip 0.1mm}c@{\hskip 0.1mm}c@{}}
&
\includegraphics[width=0.12\textwidth]{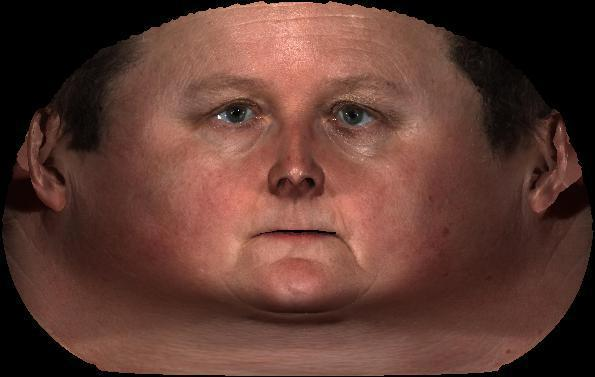} & 
\includegraphics[width=0.12\textwidth]{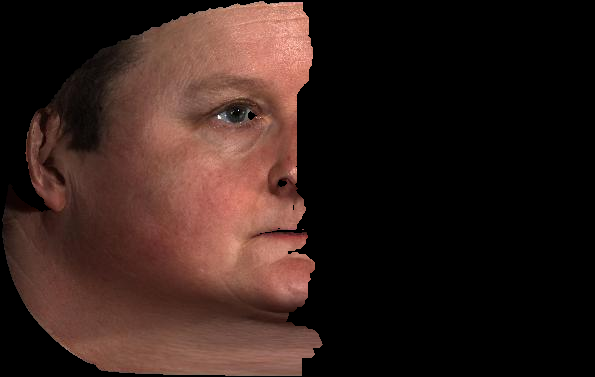} & 
\includegraphics[width=0.12\textwidth]{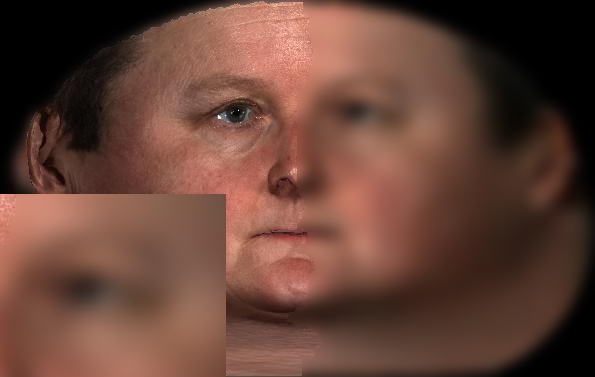} & 
\includegraphics[width=0.12\textwidth]{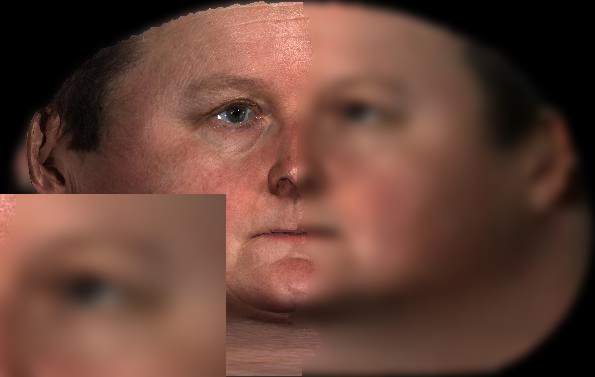} & 
\includegraphics[width=0.12\textwidth]{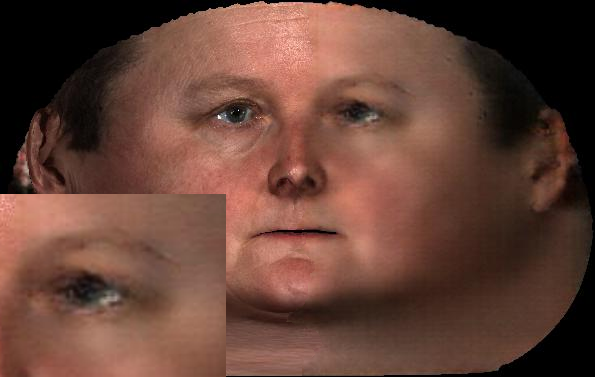} & 
\includegraphics[width=0.12\textwidth]{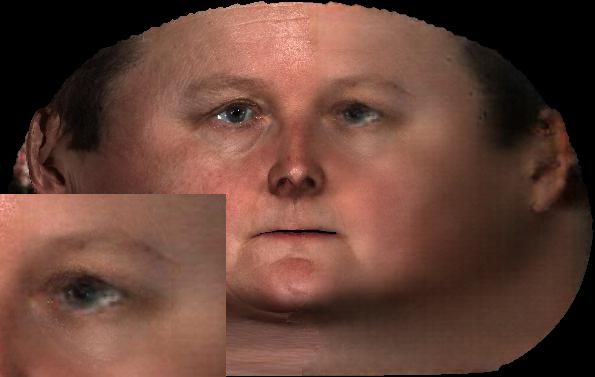} & 
\includegraphics[width=0.12\textwidth]{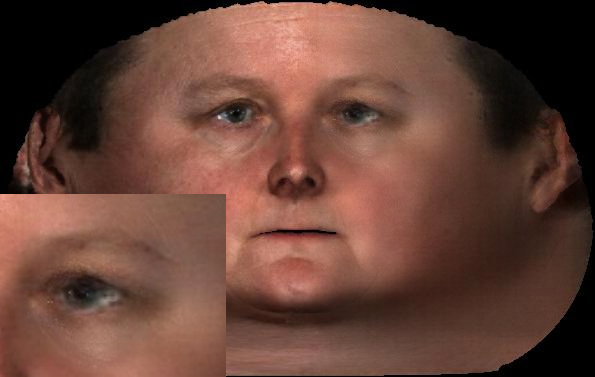} & 
\includegraphics[width=0.12\textwidth]{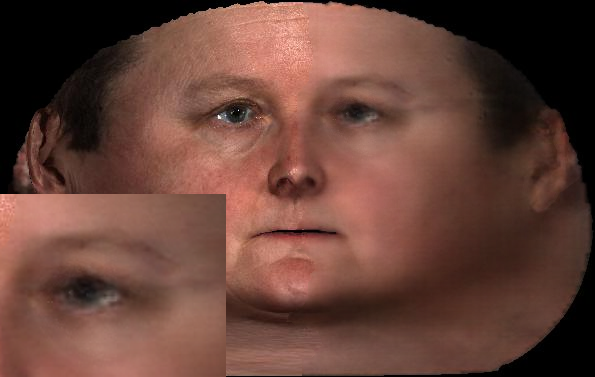} \\
&
\includegraphics[width=0.12\textwidth]{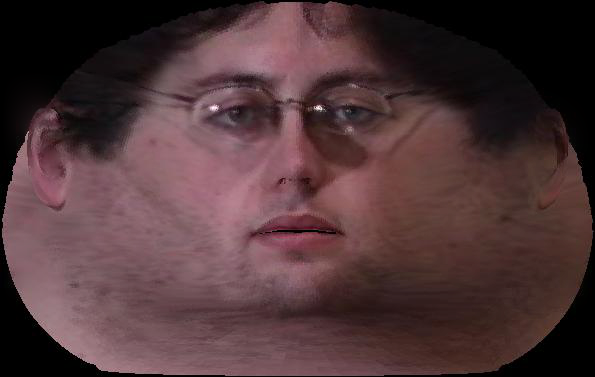} & 
\includegraphics[width=0.12\textwidth]{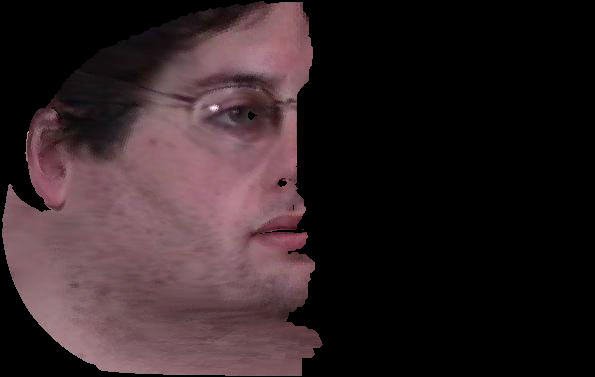} & 
\includegraphics[width=0.12\textwidth]{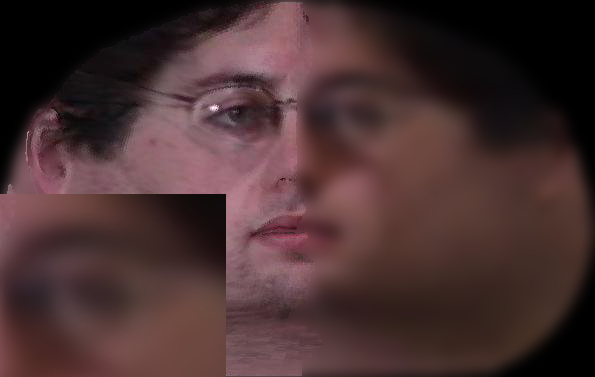} & 
\includegraphics[width=0.12\textwidth]{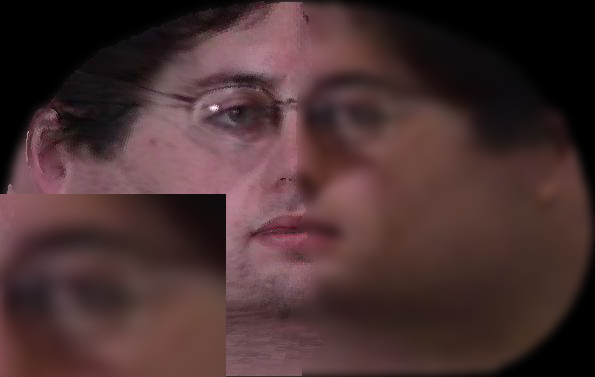} & 
\includegraphics[width=0.12\textwidth]{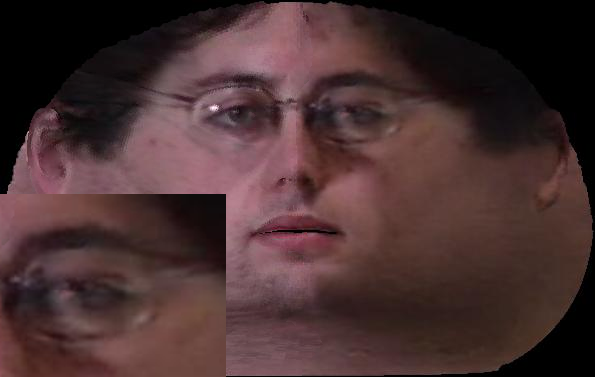} & 
\includegraphics[width=0.12\textwidth]{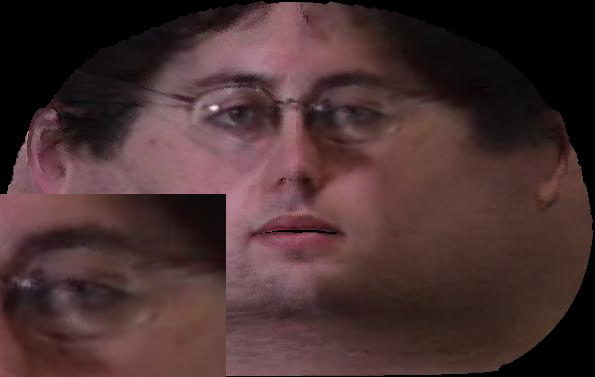} & 
\includegraphics[width=0.12\textwidth]{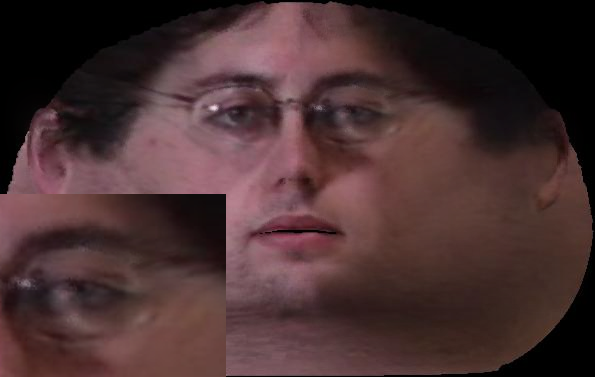} & 
\includegraphics[width=0.12\textwidth]{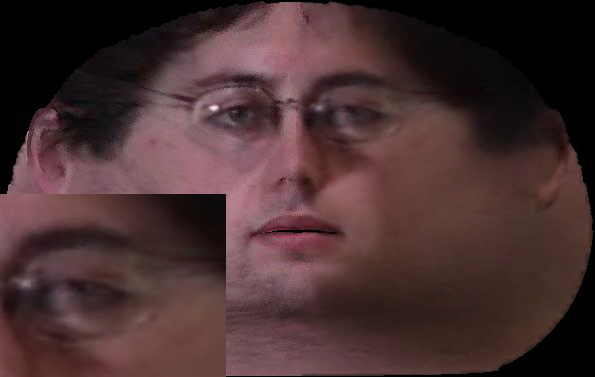} \\
&
\includegraphics[width=0.12\textwidth]{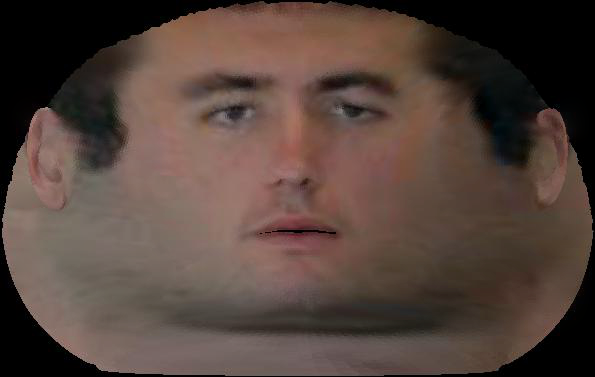} & 
\includegraphics[width=0.12\textwidth]{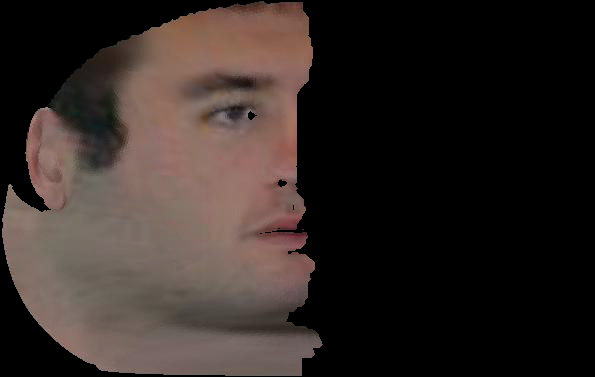} & 
\includegraphics[width=0.12\textwidth]{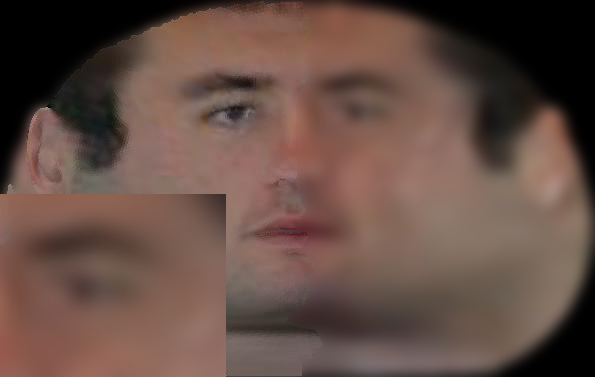} & 
\includegraphics[width=0.12\textwidth]{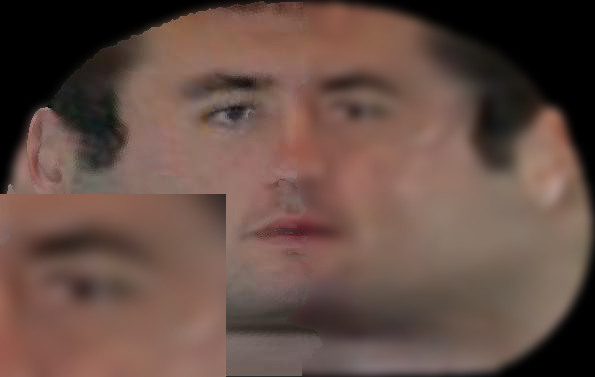} & 
\includegraphics[width=0.12\textwidth]{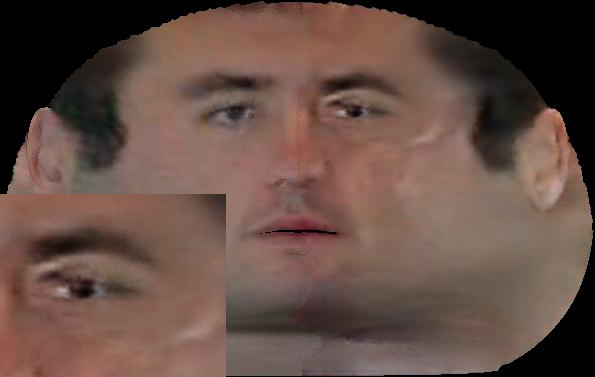} & 
\includegraphics[width=0.12\textwidth]{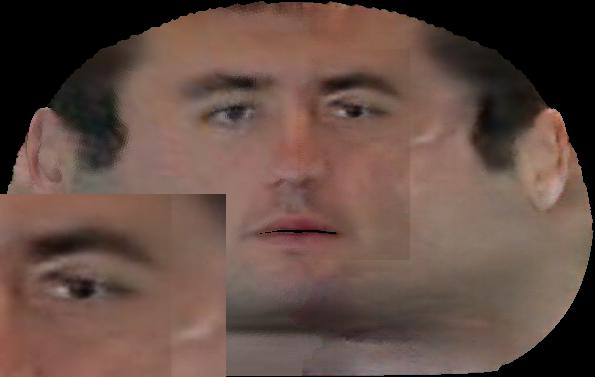} & 
\includegraphics[width=0.12\textwidth]{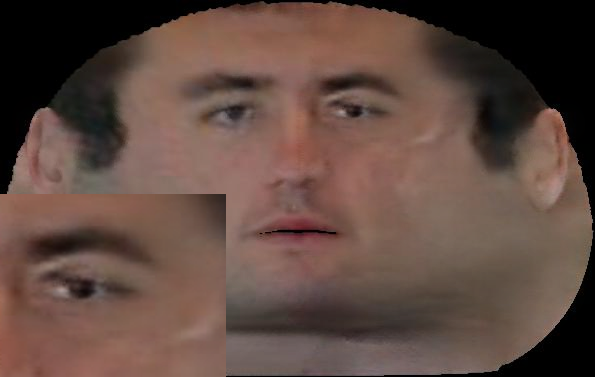} & 
\includegraphics[width=0.12\textwidth]{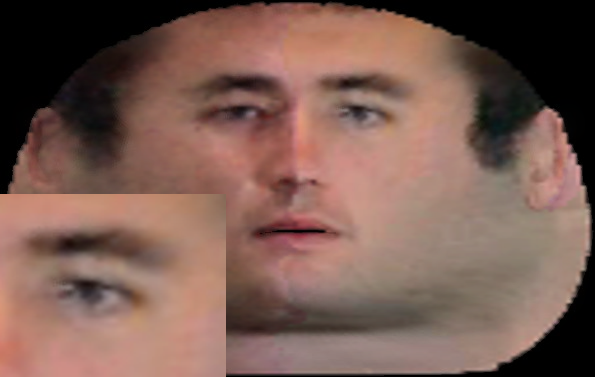} \\
\vspace{-2mm}
& GT & Masked UV  & (a) & (b) & (c) & (d) & (e) & CE~\cite{pathak2016context}\\
\end{tabular}
\end{center}
\vspace{-2mm}
\caption{UV Completion results for profile faces on UVDB under different settings of our model and Context Encoder~\cite{pathak2016context}. (a): $L_{gen}$. (b): flip + $L_{gen}$. (c): flip + $L_{gen}+ L_{adv\_g}$. (d) flip + $L_{gen}+ L_{adv\_g} + L_{adv\_l}$. (e) UV-GAN. The proposed UV-GAN shows the most realistic and plausible completed content.}
\label{fig:UVcompletionval}
\end{figure*}

\begin{table}[h!]
\small
\begin{center}
\begin{tabular}{c|c|c|c|c|c|c}
\hline
                  UVDB   & (a) & (b) & (c) & (d) & (e) & CE \\
\hline
\multirow{2}{*} {3dMD}   &  25.8  & 26.3  & 25.2  &25.7  & 26.5  & 25.1\\
                         &  0.889 & 0.895 & 0.879 &0.886 & 0.898 & 0.856\\
\hline
\multirow{2}{*} {MultiPIE}  &  25.2  & 25.7  & 24.6  &25.2  &25.8   & 24.5\\
                            &  0.881 & 0.885 & 0.865 &0.873 &0.886  & 0.842\\
\hline
\multirow{2}{*} {WildUV} & 22.3  & 22.8   & 22.0  &22.5  & 22.9  & 21.6\\
                         & 0.872 & 0.876  & 0.861 &0.868 & 0.887 & 0.840\\
\hline
\end{tabular}
\end{center}
\vspace{-2mm}
\caption{Quantitative evaluations of profile UV completion under different settings (as in Fig.~\ref{fig:UVcompletionval}) of our model and Context Encoder~\cite{pathak2016context}. For each subset, the first row is PSNR value (dB) and the second row is SSIM value.}\label{table:UV_Completionval}
\end{table}

We first conduct the ablation study of the proposed UV-GAN under different settings and also compare with Context Encoder~\cite{pathak2016context}. Fig.~\ref{fig:UVcompletionval} shows UV completion results for profile faces on the UVDB. We have also zoomed in on the occluded eye to more clearly reveal the sharpness and authenticity of the recovery result by our combined loss. With only reconstruction loss $L_{gen}$, the completed UV maps are smooth and blurry. When adding flipped faces in training, the model converges faster and leads to better results (Tab.~\ref{table:UV_Completionval}), but the completed images are still blurry. With global and local adversarial losses ($L_{adv\_g} + L_{adv\_l}$), the UV completion results look much more visually realistic and coherent globally and locally. 
Note that incorporating global adversarial loss slightly decreases PSNR and SSIM values (see Tab.~\ref{table:UV_Completionval} (c)), which are observed similarly in generative face completion~\cite{li2017generative}. Notwithstanding, when coupled with local and identity-preserving modules, such effects have been mitigated (see Tab.~\ref{table:UV_Completionval} (d) \& (e)). 
The closest to our method is the CE method proposed in~\cite{pathak2016context}. For a fair comparison, we have trained the CE method with our data. As can be seen, the proposed UV-GAN performs consistently better than the CE model both qualitatively and quantitatively. 

\begin{figure*}[h!]
\begin{center}
\begin{tabular} {@{}c@{\hskip 0.1mm}c@{\hskip 0.1mm}c@{\hskip 0.1mm}c@{\hskip 0.1mm}c@{\hskip 0.1mm}c@{\hskip 0.1mm}c@{\hskip 0.1mm}c@{\hskip 1mm}c@{}}
&
\includegraphics[width=0.12\textwidth]{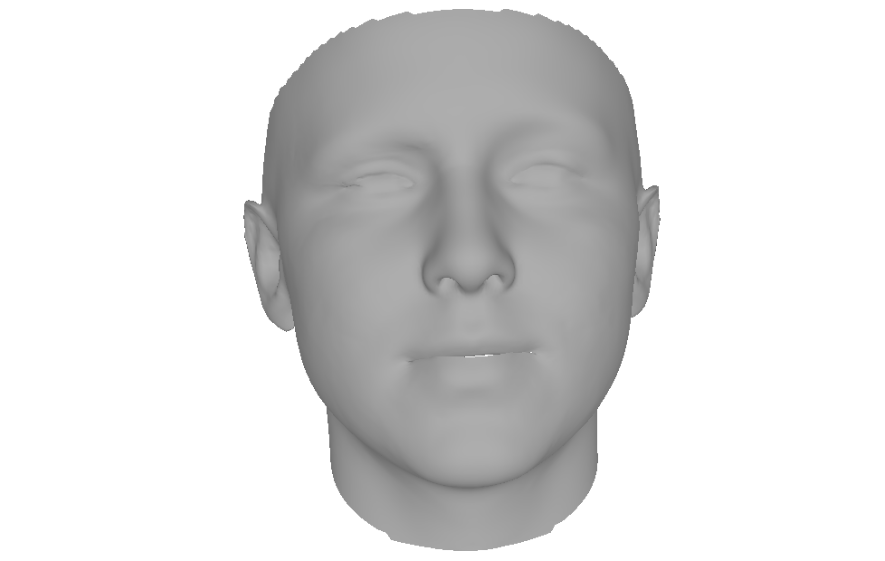} & 
\includegraphics[width=0.12\textwidth]{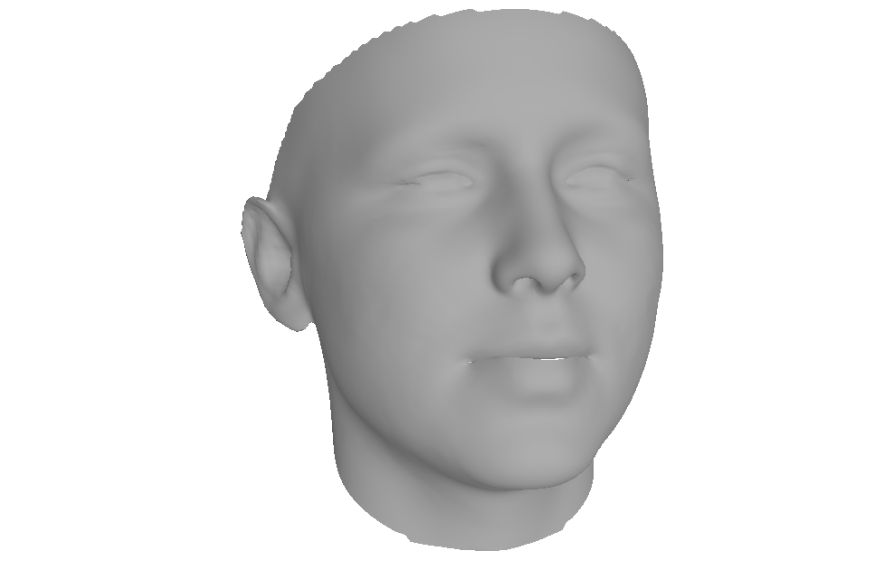} & 
\includegraphics[width=0.12\textwidth]{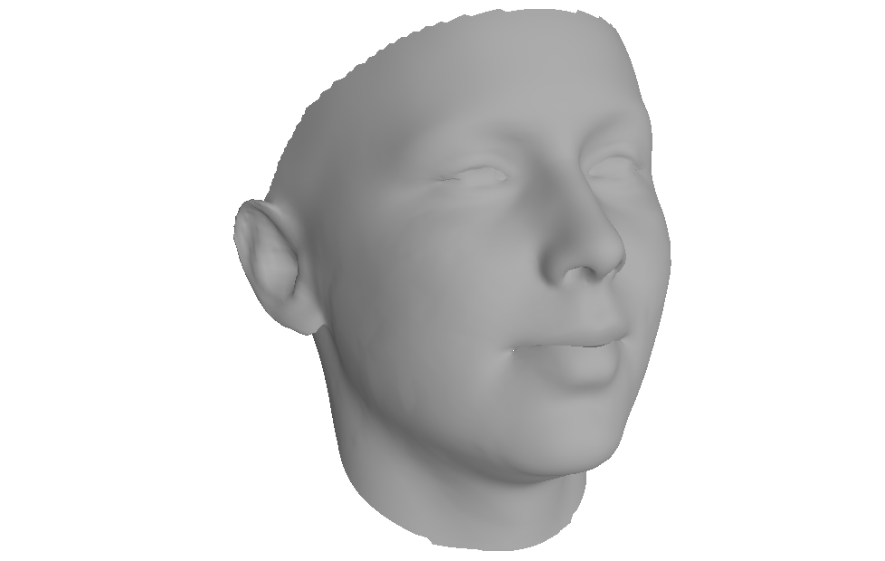} & 
\includegraphics[width=0.12\textwidth]{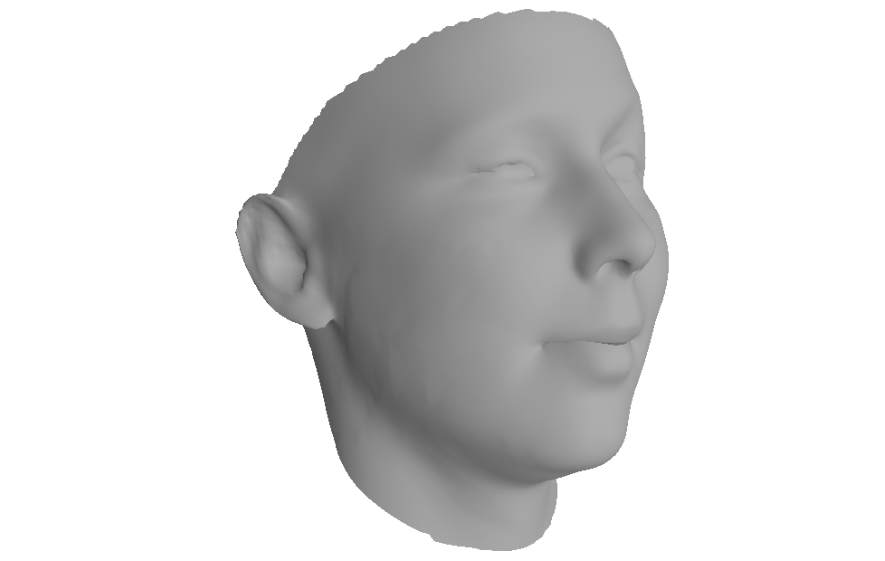} & 
\includegraphics[width=0.12\textwidth]{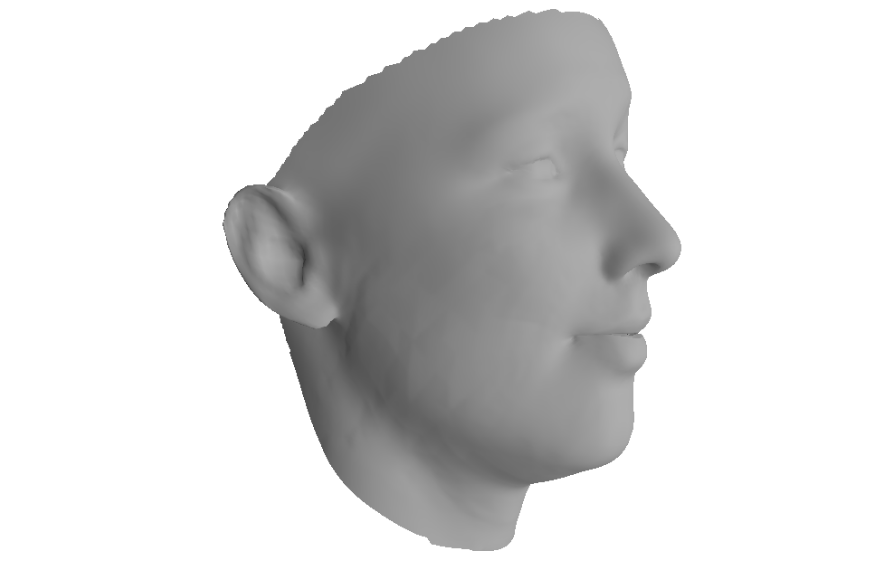} & 
\includegraphics[width=0.12\textwidth]{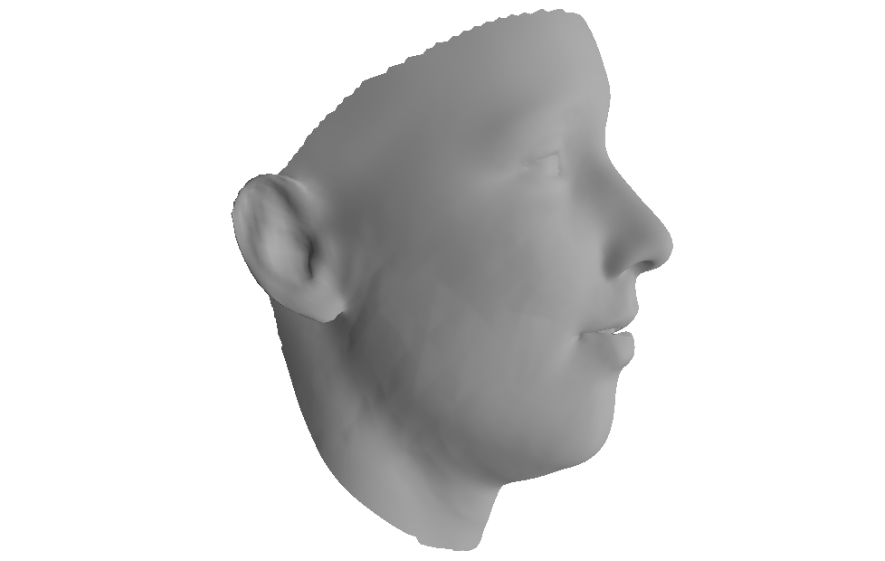} & 
\includegraphics[width=0.12\textwidth]{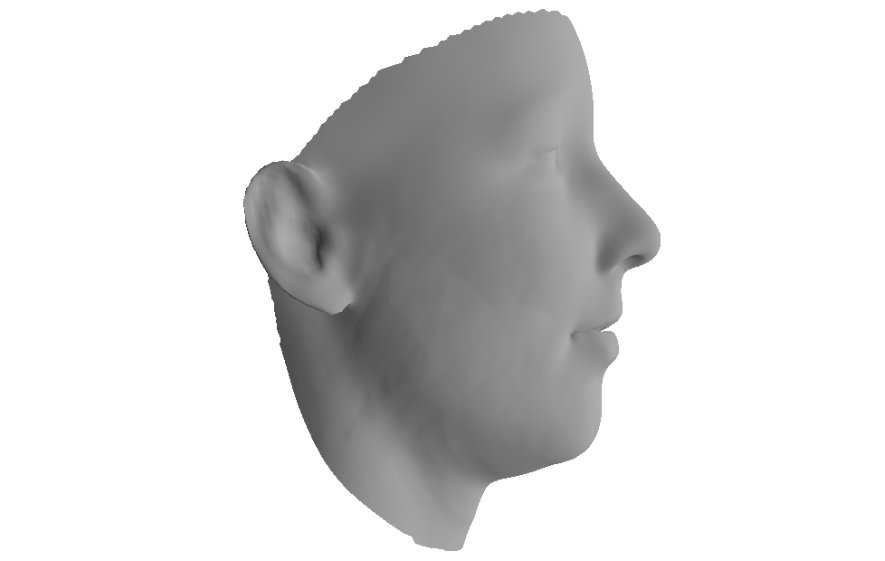} & 
Ground Truth  \\
&
\includegraphics[width=0.12\textwidth]{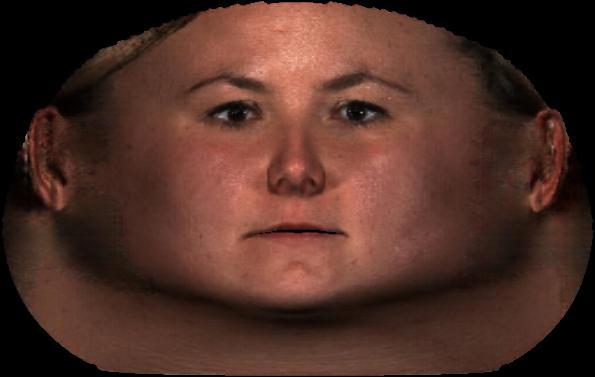} & 
\includegraphics[width=0.12\textwidth]{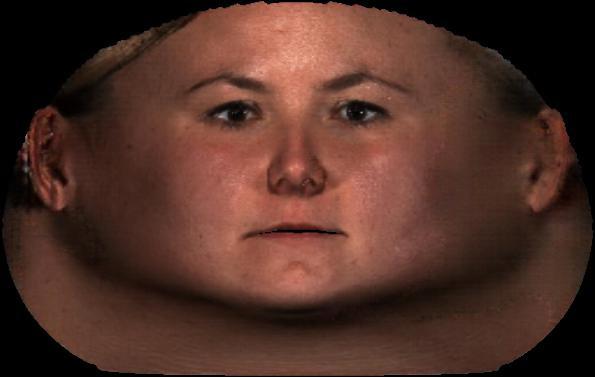} & 
\includegraphics[width=0.12\textwidth]{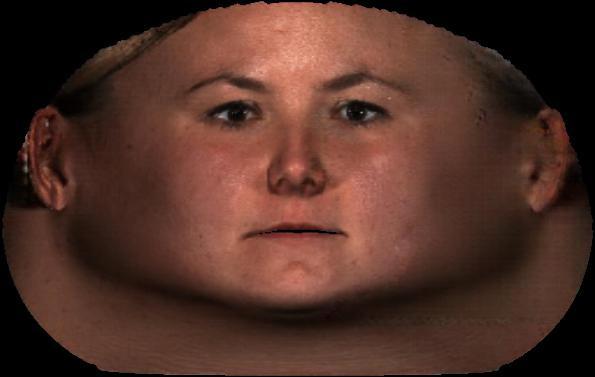} & 
\includegraphics[width=0.12\textwidth]{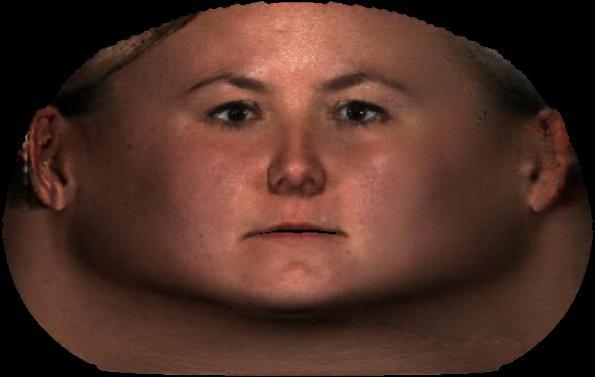} & 
\includegraphics[width=0.12\textwidth]{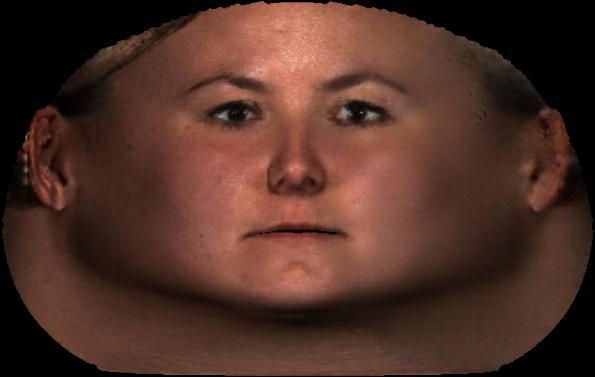} & 
\includegraphics[width=0.12\textwidth]{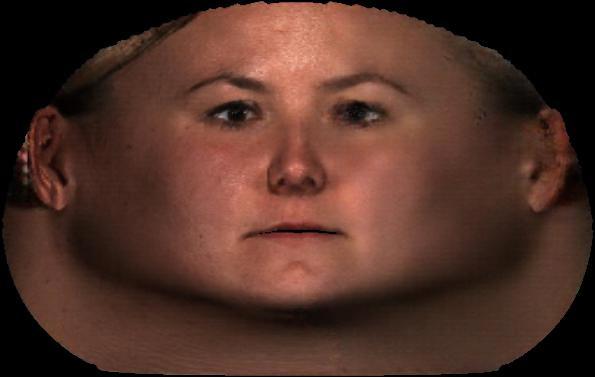} & 
\includegraphics[width=0.12\textwidth]{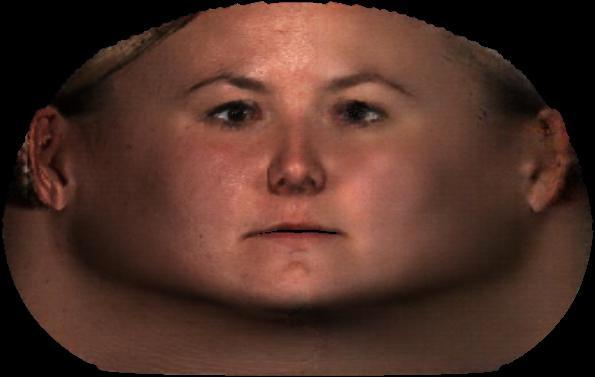} & 
\includegraphics[width=0.12\textwidth]{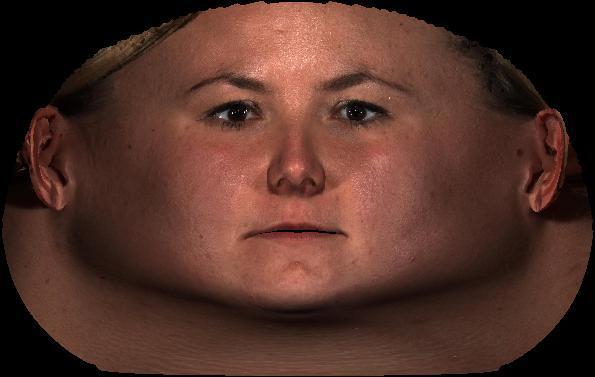}   \\
&
\includegraphics[width=0.12\textwidth]{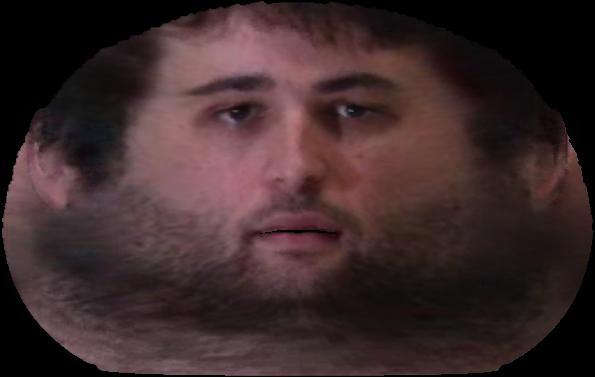} & 
\includegraphics[width=0.12\textwidth]{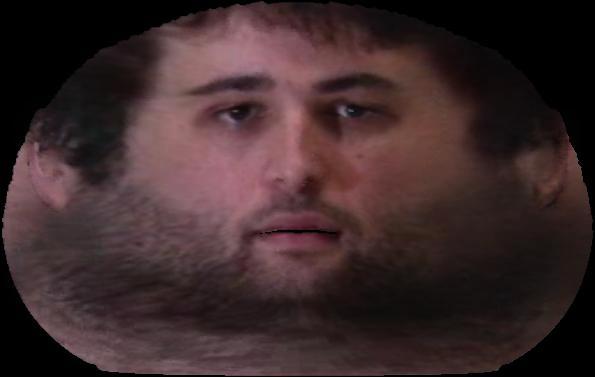} & 
\includegraphics[width=0.12\textwidth]{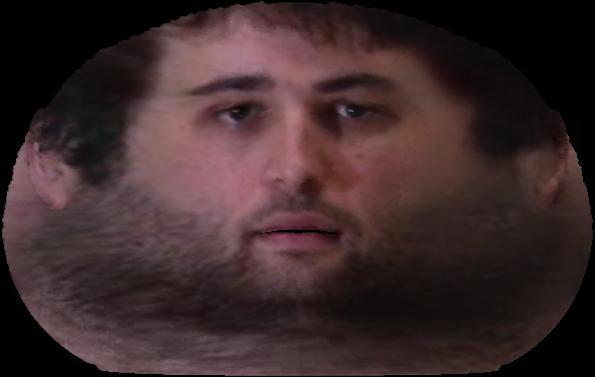} & 
\includegraphics[width=0.12\textwidth]{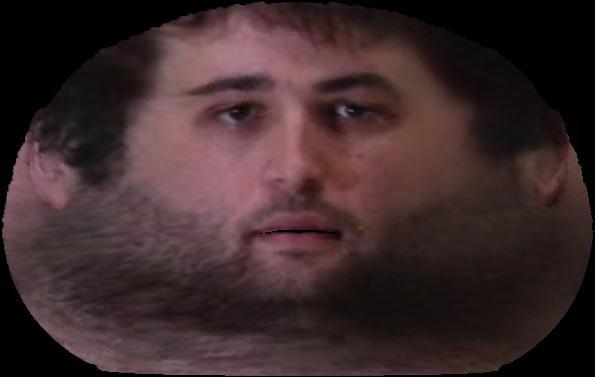} & 
\includegraphics[width=0.12\textwidth]{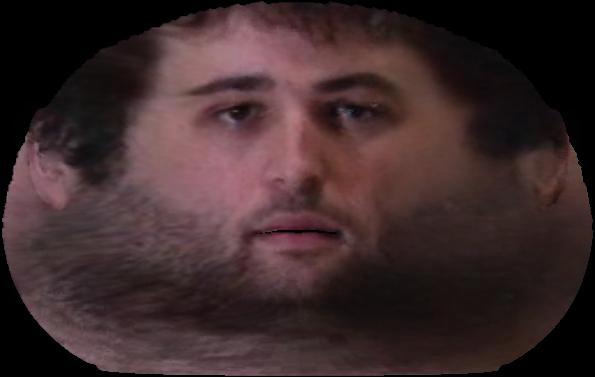} & 
\includegraphics[width=0.12\textwidth]{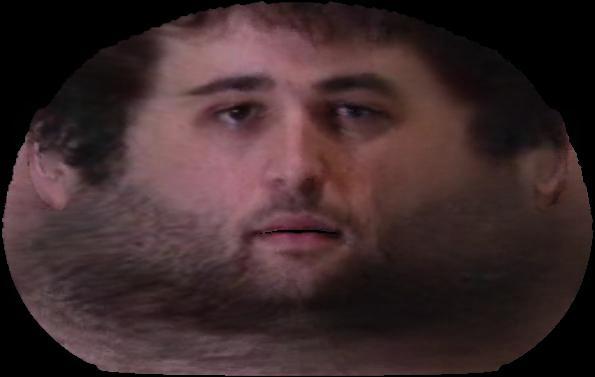} & 
\includegraphics[width=0.12\textwidth]{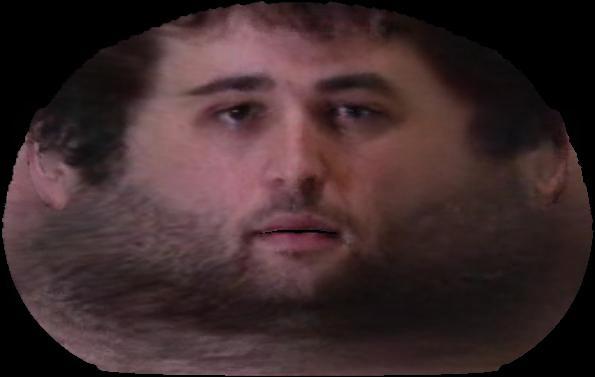} & 
\includegraphics[width=0.12\textwidth]{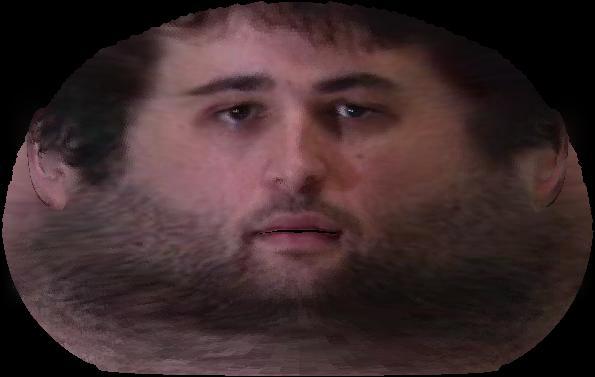} \\
&
\includegraphics[width=0.12\textwidth]{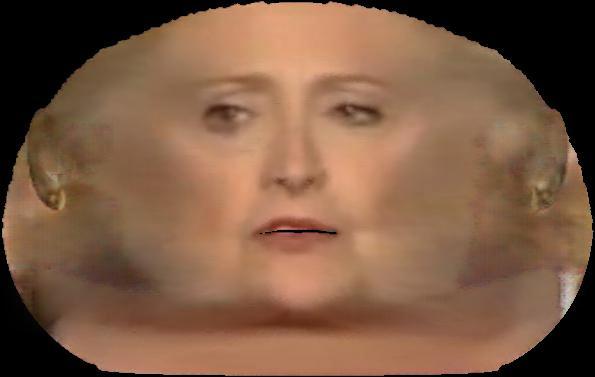} & 
\includegraphics[width=0.12\textwidth]{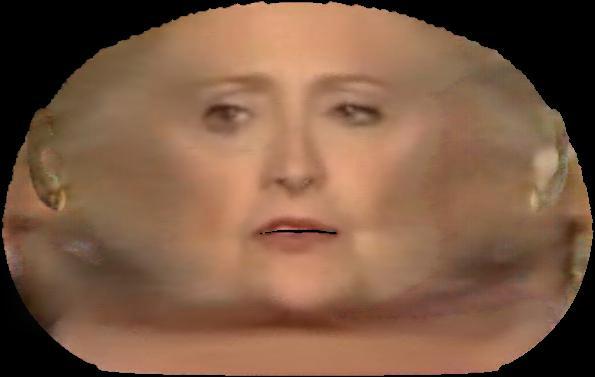} & 
\includegraphics[width=0.12\textwidth]{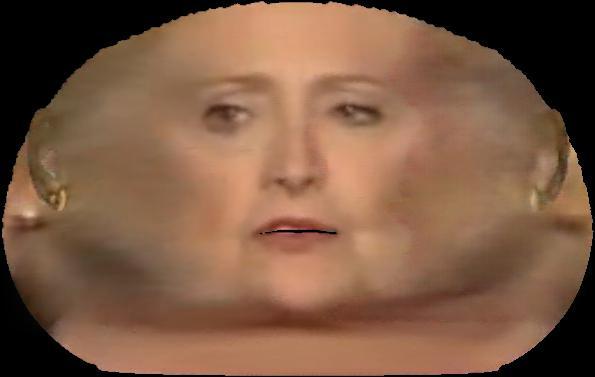} & 
\includegraphics[width=0.12\textwidth]{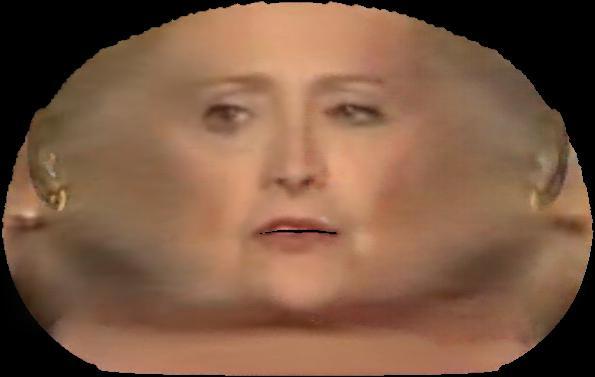} & 
\includegraphics[width=0.12\textwidth]{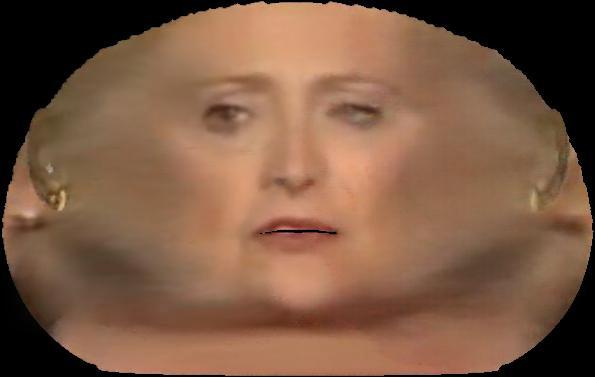} & 
\includegraphics[width=0.12\textwidth]{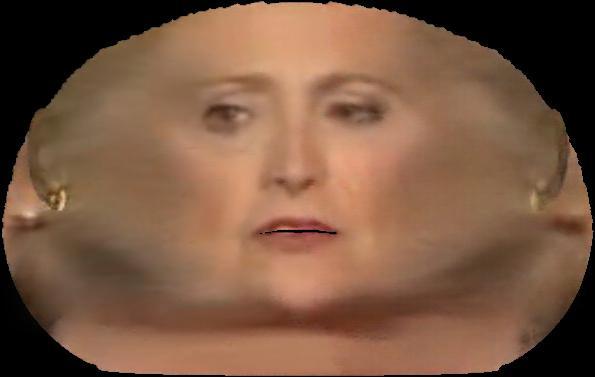} &
\includegraphics[width=0.12\textwidth]{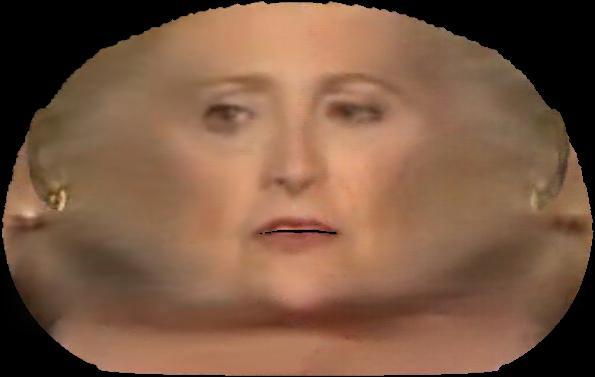} & 
\includegraphics[width=0.12\textwidth]{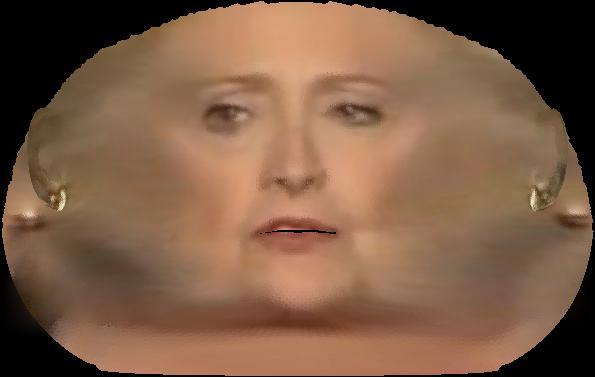} \\
\end{tabular}
\end{center}
\vspace{-2mm}
\caption{UV completion results under view changes on UVDB. Please zoom in to check the completion results on the self-occluded facial parts (\eg eye regions).}
\label{fig:UVcompletion7views}
\end{figure*}

\begin{table}[h!]
\small
\begin{center}
\begin{tabular}{c|c|c|c|c|c}
\hline
UVDB & Metric & $0^{\circ}$ & $\pm30^{\circ}$  & $\pm60^{\circ}$ & $\pm90^{\circ}$ \\
\hline
\multirow{2}{*} {3dMD} & PSNR & 28.3  & 27.9  & 27.0  & 26.5  \\
                       & SSIM & 0.920 & 0.907 & 0.903 & 0.898 \\
\hline
\multirow{2}{*} {MPIE} & PSNR & 27.5  & 26.7  & 26.5  & 25.8 \\
                       & SSIM & 0.912 & 0.903 & 0.899 & 0.886 \\
\hline
\multirow{2}{*} {WildUV} & PSNR & 26.3  & 25.2  & 24.5  & 22.9  \\
                         & SSIM & 0.908 & 0.902 & 0.894 & 0.887 \\
\hline
\end{tabular}
\end{center}
\vspace{-2mm}
\caption{Quantitative evaluations of UV-GAN under view changes.}
\label{table:UV_Completion7views}
\end{table}

Fig.~\ref{fig:UVcompletion7views} shows the performance of the proposed UV-GAN model under different poses from $0^{\circ}$ to $90^{\circ}$ in intervals of $15^{\circ}$ on the UVDB. The completion results on the controlled and in-the-wild faces are both visually realistic and consistent. In the profile case, we find that the only imperfection is the slight blurriness on the self-occluded eyes, which indicates that the proposed UV-GAN can successfully complete the UV maps in side views. In Tab.~\ref{table:UV_Completion7views}, we observe that the completion quality deteriorates as the yaw angle increases because the self-occluded regions are enlarging. Moreover, the completion results degrade as the data change from highly controlled 3dMD data to totally in-the-wild data, which indicates that UV completion in-the-wild is more challenging perhaps due to low resolution and occlusions. 

\subsection{Pose-invariant Face Recognition}

\begin{table*}[h!]
\small
\begin{center}
\begin{tabular}{c|c|ccc|ccc|ccc}
\hline
Method & LFW & 0$^\circ$- 0$^\circ$ & 0$^\circ$- 45$^\circ$ & 0$^\circ$- 90$^\circ$ & 45$^\circ$- 0$^\circ$ & 45$^\circ$- 45$^\circ$ & 45$^\circ$- 90$^\circ$ & 90$^\circ$- 0$^\circ$ & 90$^\circ$- 45$^\circ$ & 90$^\circ$- 90$^\circ$ \\
\hline
CASIA-center  &99.10& 0.8128 & 0.8001 & 0.7018 & 0.8088 & 0.8199 & 0.7328 & 0.7081 & 0.7332 & 0.7331 \\
CASIA-sphere  &99.27& 0.8010 & 0.7678 & 0.6211 & 0.7784 & 0.7934 & 0.6708 & 0.6262 & 0.6749 & 0.6957 \\
CASIA-sm      &99.02& 0.8158 & 0.7878 & 0.6483 & 0.7946 & 0.8135 & 0.6955 & 0.6515 & 0.7015 & 0.7233 \\
CASIA-sm-augUV&99.22& 0.8342 & 0.8182 & 0.7237 & 0.8256 & 0.8404 & 0.7582 & 0.7302 & 0.7597 & 0.7682 \\
\hline 
VGG2-sm       &99.35& 0.8397 & 0.8262 & \textbf{0.7325} & 0.8319 & 0.8486 & \textbf{0.7672} & \textbf{0.7386} & \textbf{0.7704} & \textbf{0.7805} \\
\hline 
MS1M-sm       &\textbf{99.60}& \textbf{0.8605} & \textbf{0.8427} & 0.6693 & \textbf{0.8486} & \textbf{0.8572} & 0.6993 & 0.6718 & 0.7050 & 0.7172 \\
\hline
\end{tabular}
\end{center}
\vspace{-2mm}
\caption{Face probing across poses on VGG2 test set. Accuracy on LFW~\cite{huang2007labeled} is also put in the left as a performance reference. Cosine similarity scores are evaluated across pose templates. A higher value is better.}\label{table:VGG2}
\end{table*}

\begin{table}[h!]
\small
\begin{center}
\begin{tabular}{c|c|c}
\hline
Method & Frontal-Frontal & Frontal-Profile \\
\hline
Human                                               &96.24 $\pm$ 0.67 & 94.57 $\pm$ 1.10\\
\hline
Sengupta \etal~\cite{sengupta2016frontal}           &96.40 $\pm$ 0.69 & 84.91 $\pm$ 1.82\\
Sankarana \etal~\cite{sankaranarayanan2016triplet}  &96.93 $\pm$ 0.61 & 89.17 $\pm$ 2.35\\
Chen \etal~\cite{chen2016fisher}                    &98.67 $\pm$ 0.36 & 91.97 $\pm$ 1.70\\
DR-GAN~\cite{tran2017disentangled}                  &97.84 $\pm$ 0.79 & 93.41 $\pm$ 1.17\\
DR-GAN+~\cite{tran2017representation}               &97.84 $\pm$ 0.79 & 93.41 $\pm$ 1.17\\
Peng \etal~\cite{Peng_2017_ICCV}                    &98.67            & 93.76           \\
\hline
CASIA-center                                        &98.34 $\pm$ 0.44 & 87.77 $\pm$ 2.39\\
CASIA-Sphere                                        &98.64 $\pm$ 0.24 & 84.39 $\pm$ 2.59\\
CASIA-sm                                            &98.59 $\pm$ 0.21 & 87.74 $\pm$ 1.07\\
CASIA-sm-aug1                                       &98.25 $\pm$ 0.42 & 90.14 $\pm$ 1.53\\
\hline
CASIA-sm-augUV                                      &98.83 $\pm$ 0.27 & 93.09 $\pm$ 1.72\\
-Profile2Frontal                                    &-               & 93.55  $\pm$ 1.67\\
-Frontal2Profile                                    &-               & 93.72  $\pm$ 1.59\\
-Template2Template                                  &-               & {\bf 94.05}$\pm$ 1.73\\
\hline
VGG2-sm                                             &99.17 $\pm$ 0.11 & 93.40 $\pm$ 1.64\\
MS1M-sm                                             &{\bf 99.59} $\pm$ 0.13 & 87.11 $\pm$ 1.47\\
\hline
Shape Only                                          &67.49 $\pm$ 2.04 & 62.26 $\pm$ 2.57 \\
\hline
\end{tabular}
\end{center}
\vspace{-2mm}
\caption{Verification accuracy($\%$) comparison on CFP dataset.}
\vspace{-5pt}
\label{table:CFP}
\end{table}
\begin{figure*}[h!]
  \centering
  \subfigure[Profile Face UV Completion]{
    \label{fig:subfig:CFP_profile} 
    \includegraphics[width=0.9\textwidth]{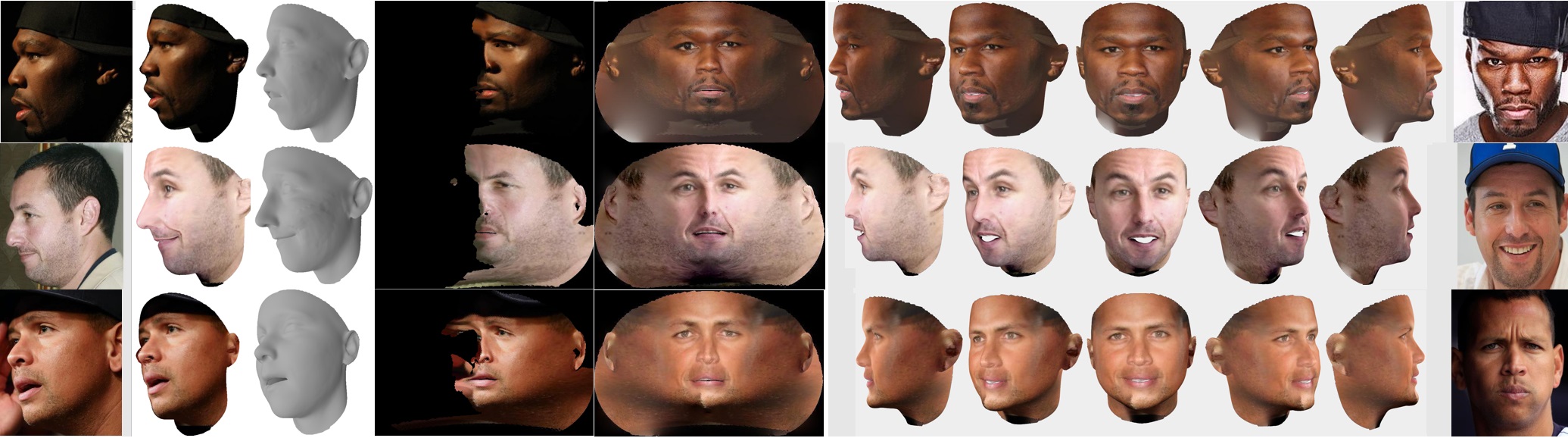}}
 \subfigure[Frontal Face UV Completion]{
    \label{fig:subfig:CFP_fontal} 
    \includegraphics[width=0.9\textwidth]{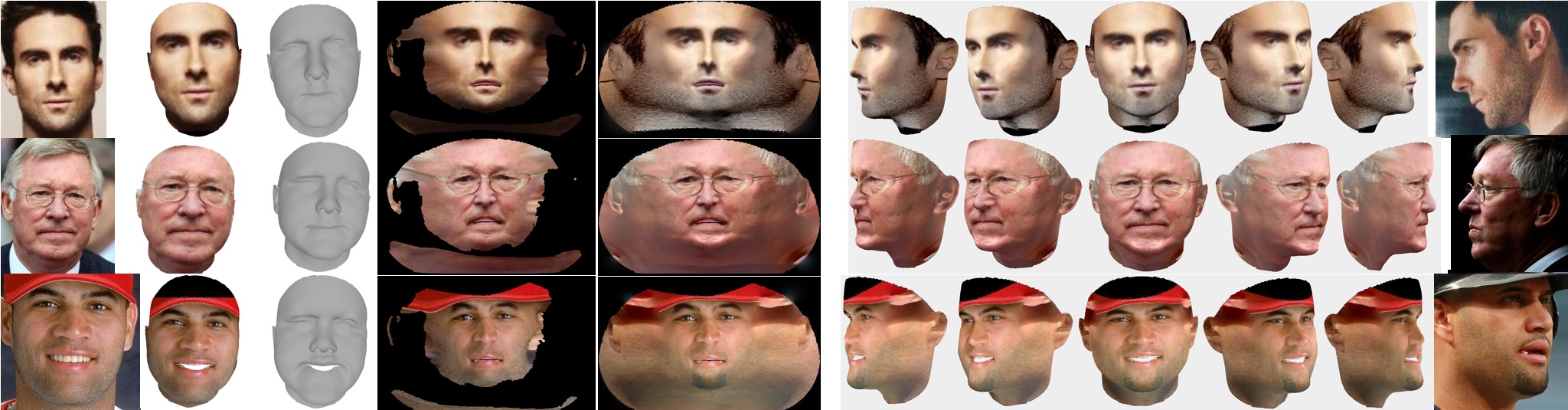}}
 \vspace{-2mm}
 \caption{Profile face and frontal face UV Completion on CFP dataset. From left to right: 2D face images, 3D face fitting results, 3D face shapes, self-occluded UV maps, UV completion results by UV-GAN, 3D face synthesis of five views, and ground truth of the frontal/profile faces. The proposed UV-GAN can generate realistic, coherent and identity-preserved UV maps for in-the-wild profile and frontal faces.}
\label{fig:CFP_completion}
\end{figure*}

We pre-process all the face images by applying the face detection~\cite{zhang2016joint} and a 3DMM fitting. After that, face images are cropped to $112\times112$. To study the impacts of training data and loss function on recognition, we use the same network architecture in all our experiments (ResNet-27\footnote{https://github.com/ydwen/caffe-face}~\cite{wen2016discriminative}) without any bells and whistles. That is, we only vary the training data and the loss functions, which include centre loss~\cite{wen2016discriminative}, sphere loss~\cite{liu2017sphereface} and softmax loss. 

We abbreviate each training strategy as [dataset]-[loss]-[aug]. For example, CASIA-sm means that we have trained the network using CASIA dataset with the soft-max loss (denoted by sm), while CASIA-sphere means that we have used the sphere loss. CASIA-sm-aug1 refers to training the network in the CASIA dataset using the softmax loss (sm) and the augmentation method proposed in~\cite{masi2016we}. CASIA-sm-augUV refers to training using the same dataset and loss as above but augmenting the data using UV-GAN. Our augmentation method is as follows: we create 300 face images per identity (similar to Fig.~\ref{fig:CFP_completion}). For frontal faces in the training data, we generate non-frontal views, which preserve accurate facial details that are critical for identity discrimination. For non-frontal faces in the training data, we only increase the pose angle to generate non-frontal faces with larger self-occluded area. 

{\bf VGG2} This experiment aims at assessing the impact of pose variations (frontal, three-quarter and profile views) on template matching. In~\cite{cao2017vggface2}, six templates of each subject are divided into two sets, A and B, so that each set contains templates of three different poses. And a $3\times3$ similarity matrix is constructed between the two sets. Each template is represented by a single vector computed by averaging the face descriptors of each face in the set. The similarity between templates is then computed as the cosine distance between the vectors representing each template.

In Tab.~\ref{table:VGG2}, we show the similarity matrix averaged over 368 subjects. Each row corresponds to the three pose templates from A versus the three pose templates from B. We find that (1) the similarity drops greatly when probing pairs of differing poses, \eg, front-to-three-quarter and front-to-profile, implying that recognition across poses is a challenging problem; (2) recognition results are better when matching pairs with smaller pose gaps, which means that minimising pose gaps during testing, \eg, face frontalisation or frontal face rotation, is highly beneficial;
(3) recognition results are better when the training data contain large pose variations (\eg VGG2). That is, pose-invariant feature embedding can be directly learned from such data without any special tricks; (4) by simply increasing the number of identities (\eg using MS1M~\cite{guo2016ms}) without having large pose variations in the training data, it is not possible to learn a pose-invariant embedding; (5) training on the proposed CASIA-augUV dataset is more beneficial for learning a pose-invariant embedding than training on the CASIA dataset; (6) the performance of a model trained on the CASIA-augUV approaches the performance of a model trained in VGG2 dataset, which is 6 times larger than CASIA. Hence, it is evident that by incorporating larger pose variations in the training set using the proposed UV-GAN improves the performance. 

{\bf CFP} The Celebrities in Frontal-Profile (CFP) dataset~\cite{sengupta2016frontal} focuses on extreme pose face verification. The reported human performance is $96.24\%$ on the frontal-frontal protocol and $94.57\%$ on the frontal-profile protocol, which shows the challenge in recognising profile views. 

In Tab.~\ref{table:CFP}, we compare various training strategies with state-of-the-art methods. For the frontal-frontal protocol, the model MS1M-sm obtains the best verification accuracy with $99.59\%$, and the model VGG2-sm ranks second with an accuracy of $99.17\%$, which indicates that MS1M and VGG2 contain larger variations than CASIA dataset. 
Nevertheless, it is evident that our data augmentation on CASIA helps to improve the performance from $98.59\%$ to $98.83\%$. By contrast, data augmentation by~\cite{masi2016we} (CASIA-sm-aug1) only gives an accuracy of $98.25\%$. The method in~\cite{masi2016we} augments the data by introducing perturbation on the parameters of the 3D shape that correspond to identity. We believe that this is not an optimal strategy, since these parameters contain important information regarding identity. To support our claim, the last row of Tab.~\ref{table:CFP} shows the performance of the 157-d identity parameters from our 3D fitting results. The recognition accuracy is $67.49\%$ which demonstrates that even though the shape is quite low-dimensional and only estimated by 3DMM fitting it indeed contains information regarding identity. 

For the frontal-profile protocol on the CFP dataset, pose augmentation during training is very effective in learning pose-invariant feature embedding. Our augmentation improves the accuracy from $87.74\%$ (CASIA-sm) to $93.09\%$ (CASIA-sm-augUV) and even approaches the performance of the VGG2-sm ($93.40\%$), which is trained on a much larger dataset than CASIA (see Tab.~\ref{table:CFP}). To further improve the performance, we use our UV-GAN to synthesise frontal faces from profile faces and vice versa to producing matching pairs in the testing set. We present some illustrations of the completed UV maps for profile and frontal faces in Fig.~\ref{fig:CFP_completion}. From the completed UV maps and 3D face shapes, we can synthesise the face in arbitrary poses, \eg frontal and profile faces. 

By synthesising frontal faces from profile faces (Profile2Frontal in Tab.~\ref{table:CFP} and Fig.~\ref{fig:subfig:CFP_profile}) during testing and matching them, the accuracy improves by $0.46\%$. Similarly, synthesising profile faces from frontal faces (Frontal2Profile in Tab.~\ref{table:CFP} and Fig.~\ref{fig:subfig:CFP_fontal}) leads to slightly better results, an improvement of $0.63\%$.  Since our UV-GAN can generate faces with arbitrary poses from any given face, we can easily translate the pair-wise face verification problem into a more robust template verification problem. In our experiments, we synthesise frontal faces from profile faces and profile faces from frontal faces at the same time, which are denoted by Template2Template in Tab.~\ref{table:CFP}. We have used a view interpolation of $15^\circ$ to generate two templates. After that, we used the generated template feature centres to conduct verification, where the accuracy attains a value of $94.05\%$, $0.29\%$ higher than the state-of-art method proposed by Peng~\etal~\cite{Peng_2017_ICCV}. 

\section{Conclusions}

In this paper, an understudied computer vision problem, that of completion of facial UV maps that have been produced by fitting 3D face models in images. To this end, we collected a large-scale dataset of completed facial UV maps. Then, we employ global and local adversarial networks to learn identity-preserved UV completion. When we attach the completed UV map to the fitted 3D mesh, we can get faces with arbitrary poses, which can increase pose variations during training of a face recognition model and decrease pose discrepancy during testing, which lead to better performance. Experiments on both controlled and in-the-wild UV datasets confirm the effectiveness of the proposed UV completion method. The proposed method also obtains state-of-the-art verification accuracy under the CFP frontal-profile evaluation protocol.

{\small
\bibliographystyle{ieee}
\bibliography{egbib}
}

\end{document}